\documentclass[graybox]{svmult}    

\usepackage{mathptmx}       
\usepackage{helvet}         
\usepackage{courier}        
\usepackage{type1cm}        
\usepackage{makeidx}         
\usepackage[pdftex]{graphicx}        
\usepackage{multicol}        
\usepackage[bottom]{footmisc}

\makeindex             

\usepackage{geometry}
\usepackage[utf8]{inputenc} 
\usepackage[T1]{fontenc}    
\usepackage{hyperref}       
\usepackage{url}            
\usepackage{booktabs}       
\usepackage{amsfonts}       
\usepackage{nicefrac}       
\usepackage{microtype}      
\usepackage{tabularx}
\usepackage{caption}
\usepackage{mdwlist}
\usepackage{amssymb}
\usepackage{diagbox}
\usepackage{wrapfig}
\usepackage{listings}
\usepackage{color}
\usepackage{xcolor}
\usepackage{nicefrac}
\usepackage{subcaption}
\usepackage{amsmath}

\usepackage{algorithm}
\usepackage{algorithmicx}
\usepackage{algpseudocode}
\usepackage{varwidth}

\usepackage{tikz}

\global\long\def\E{\mathbb{E}}

\usepackage{graphicx}

\usepackage{todonotes}
\usepackage{float}
\usepackage{comment}
\usepackage[normalem]{ulem}

\usepackage{caption}

\usepackage{titlesec} 

\usepackage{verbatim}
 \usetikzlibrary{decorations.pathreplacing}

\DeclareMathOperator*{\argmax}{arg\,max} 

\newcommand{\citep}[1]{\cite{#1}} 

\makeatletter
\newcommand{\sectionauthor}[1]{%
  {\parindent0pt\vspace*{-12pt}%
  \linespread{1.1}#1%
  \par\nobreak\vspace*{15pt}}
  \@afterheading%
}
\makeatother

\graphicspath{
{img/}
}
\begin{document}
\title*{Learning to Run challenge solutions: Adapting reinforcement learning methods for neuromusculoskeletal environments}

\author{{\L}ukasz Kidzi\'nski, Sharada Prasanna Mohanty, Carmichael Ong, Zhewei Huang, Shuchang Zhou, Anton Pechenko, Adam Stelmaszczyk, Piotr Jarosik, Mikhail Pavlov, Sergey Kolesnikov, Sergey Plis, Zhibo Chen, Zhizheng Zhang, Jiale Chen, Jun Shi, Zhuobin Zheng, Chun Yuan, Zhihui Lin, Henryk Michalewski, Piotr Miłoś, Błażej Osiński, Andrew Melnik, Malte Schilling, Helge Ritter, Sean Carroll, Jennifer Hicks, Sergey Levine, Marcel Salathé, Scott Delp}


\maketitle
\abstract{In the NIPS 2017 \textit{Learning to Run} challenge, participants were tasked with building a controller for a musculoskeletal model to make it run as fast as possible through an obstacle course. Top participants were invited to describe their algorithms. In this work, we present eight solutions that used deep reinforcement learning approaches, based on algorithms such as Deep Deterministic Policy Gradient, Proximal Policy Optimization, and Trust Region Policy Optimization. Many solutions use similar relaxations and heuristics, such as reward shaping, frame skipping, discretization of the action space, symmetry, and policy blending. However, each of the eight teams implemented different modifications of the known algorithms.}


\section{Introduction}
In the Learning to Run challenge participants were tasked to build a controller for a human musculoskeletal model, optimizing muscle activity such that the model travels as far as possible within 10 seconds \cite{kidzinski2018learningtorun}. Participants were solving a control problem with a continuous space of $41$ input and $18$ output parameters with high order relations between actuations and actions, simulating human musculoskeletal system. Expensive computational cost of the musculoskeletal simulations encouraged participants to develop new techniques tailored for this control problem.

All participants whose models traveled at least 15 meters in 10 seconds of the simulator time were invited to share their solutions in this manuscript. Nine teams agreed to contribute. The winning algorithm is published separately \cite{jaskowski2018rltorunfast}, while the remaining eight are collected in this manuscript. Each section in the reminder of this document describes an approach taken by one team. Sections are self-contained, they can be read independently, and each of them starts with an introduction summarizing the approach. For information on compositions of teams, affiliations and acknowledgments please refer to Section \ref{s:acknowledgements}. 
\section{Learning to Run with Actor-Critic Ensemble}\label{s:pku}
\sectionauthor{Zhewei Huang and Shuchang Zhou}

We introduce an Actor-Critic Ensemble (ACE) method for improving the performance of Deep Deterministic Policy Gradient (DDPG) algorithm\citep{lillicrap2015continuous,silver2014deterministic}. At inference time, our method uses a critic ensemble to select the best action from proposals of multiple actors running in parallel. By having a larger candidate set, our method can avoid actions that have fatal consequences, while staying deterministic. Using ACE, we have won the 2nd place in NIPS'17 Learning to Run competition. 

\subsection{Methods}
\subsubsection{Dooming Actions Problem of DDPG}
We found that in the \textit{Learning to Run} challenge environment legs of a fast running skeleton can easily be tripped up by obstacles. This caused the skeleton to enter an unstable state with limbs swinging and falling down after a few frames. We observed that it was almost impossible to recover from the unstable states. We call the action causing the skeleton to enter unstable state a ``dooming action''. 

\begin{figure*}[ht!]%
        \centering
        \begin{subfigure}[t]{0.45\textwidth}
                \includegraphics[width=1\textwidth]{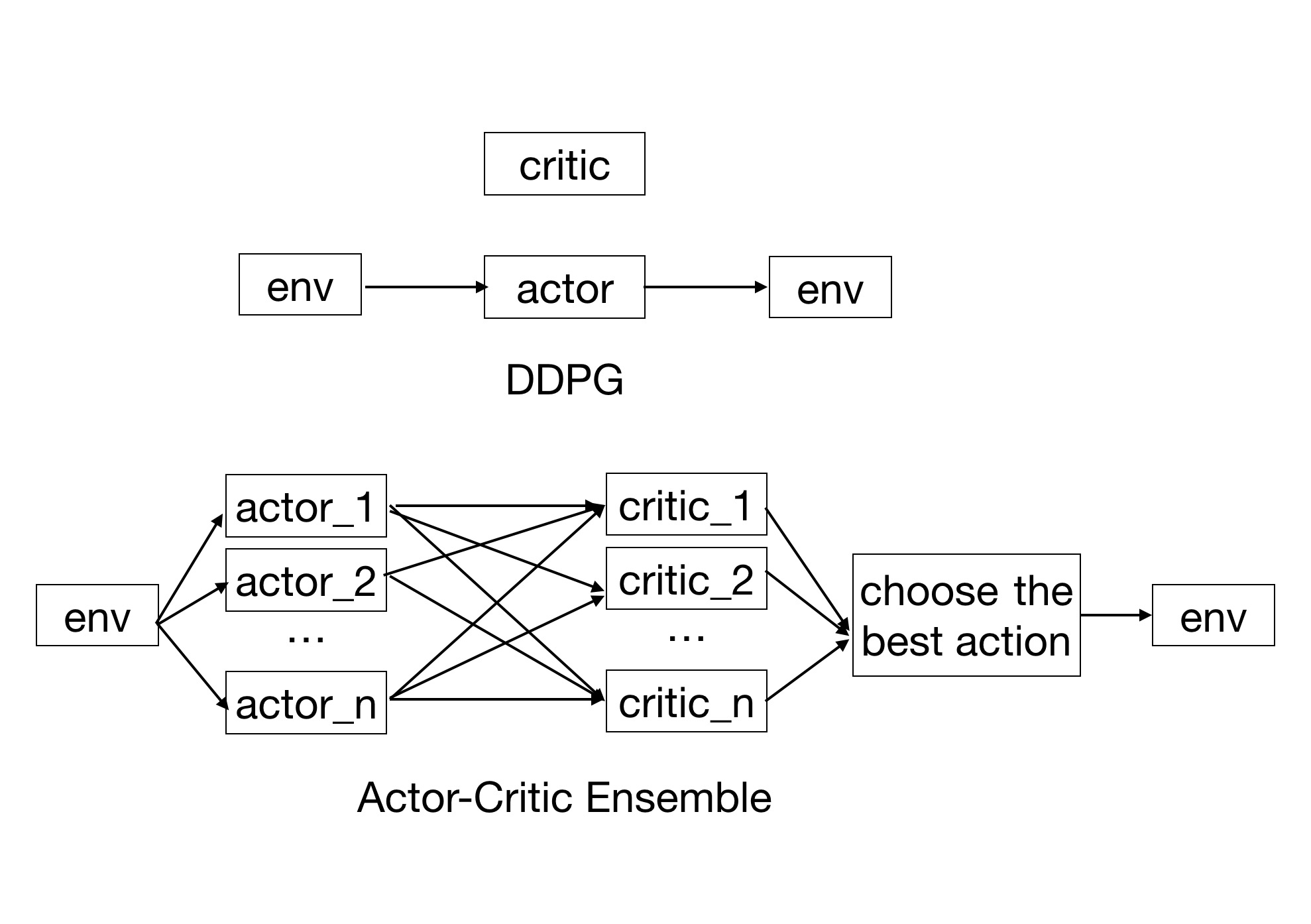}
                \label{fig:ddpg-ace}
        \end{subfigure}
        \begin{subfigure}[t]{0.45\textwidth}
                \includegraphics[width=1\textwidth]{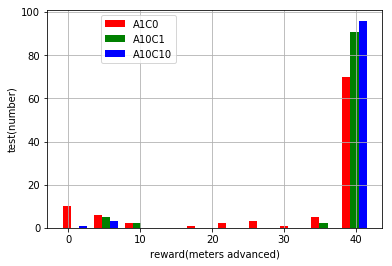}
                \label{fig:performance}
        \end{subfigure}
        \caption{Schema for DDPG and ACE}
        \label{fig:ace}      
\end{figure*}

To investigate dooming actions, we let the critic network inspect the actions at inference time. We found that most of the time, the critic could recognize dooming actions by anticipating low scores. However, as there was only one action proposed by the actor network in DDPG at every step, the dooming actions could not be avoided. This observation led us to use an actor ensemble to allow the agent to avoid dooming actions by having a critic ensemble to pick the best action from the proposed ones, as shown in Fig.~\ref{fig:ace}(a).

\subsubsection{Inference-Time ACE}
We first trained multiple actor-critic pairs separately, using the standard DDPG method. Then we built a new agent with many actor networks proposing actions at every step. Given multiple actions, a critic network was used to select the best action. The actor picked the action with the highest score in a greedy manner.

Empirically, we found that actors of heterogeneous nature, e.g. trained with different hyper-parameters, perform better than actors from different epochs of the same training setting. This was in agreement with the observations in the original work on Ensemble Learning \citep{dietterich2000ensemble}.

To further improve critic's prediction quality, we built an ensemble of critics, by picking the pairing critics of actors. We combined the outputs of the critic networks by averaging them.

\subsubsection{Training with ACE}
If we put actor networks together to train, all the actor networks are updated at every step, even if a certain action was not used. The modified Bellman equation takes form
\begin{align*}
i_{t+1} = \arg\max_j{Q(s_{t+1},\mu_j(s_{t+1}))}\\
Q(s_t, a_t) = r(s_t, a_t) + \gamma Q(s_{t+1},\mu_{i_{t+1}}({s_{t+1}})).
\end{align*}
\subsection{Experiments and results}
\subsubsection{Baseline Implementation}
We used the DDPG as our baseline. To describe the state of the agent, we collected three consecutive frames of observations from the environment. We performed feature engineering as proposed by Yongliang Qin\footnote{\url{https://github.com/ctmakro/stanford-osrl}}, enriching the observation before we feeding into the network.

As the agent was expected to run 1000 steps to finish a successful episode, we found the vanishing gradient problem (i.e. too small magnitude of the update step in the learning process) to be critical. We made several attempts to deal with this difficulty. First, we found that with the original simulation timestep, the DDPG converges slowly. In contrast, using four times larger simulation timestep, which was equivalent to changing the action only every four frames, was found to speedup convergence significantly.

We also tried unrolling DDPG as in $TD(\lambda)$ with $\lambda=4$ \citep{anonymous2018distributional}, but found it inferior to simply increasing simulation timestep. Second, we tried several activation functions and found that the activation function of Scaled Exponential Linear Units(SELU)\citep{klambauer2017self} is superior to ReLU, as shown in Fig.~\ref{fig:ablation}. SELU also outperformed Leaky ReLU, Tanh and Sigmoid.

\begin{figure*}[ht!]%
        \centering
        \begin{subfigure}[t]{0.45\textwidth}
                \includegraphics[width=1\textwidth]{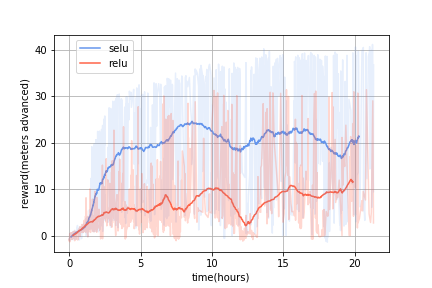}
                \label{fig:relu-selu-60}
        \end{subfigure}
        \begin{subfigure}[t]{0.45\textwidth}
                \includegraphics[width=1\textwidth]{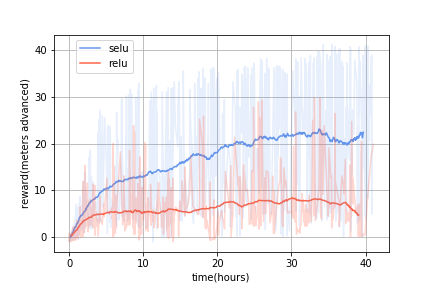}
                \label{fig:relu-selu-20}
        \end{subfigure}
        \caption{Training with different activation functions and different number of processes for generating training data, by DDPG}
        \label{fig:ablation}
\end{figure*}

\begin{table}
\caption{Hyper-parameters used in the experiments}
\begin{tabularx}{\columnwidth}{r|X}
  \toprule
  Actor network architecture & $[FC800, FC400]$, Tanh for output layer and SELU for other layers\\
  Critic network architecture & $[FC800, FC400]$, linear for output layer and SELU for other layers\\
  Actor learning rate & 3e-4 \\
  Critic learning rate & 3e-4 \\
  Batch size & 128 \\
  $\gamma$ & 0.96 \\
  replay buffer size & $2\mathrm{e}{6}$\\
                          \bottomrule
\end{tabularx}
\label{tab:hyperopt}
\end{table}

\subsubsection{ACE experiments}
For all models we used an identical architecture of actor and critic networks, with hyper-parameters listed in Table~\ref{tab:hyperopt}. Our code used for competition can be found online\footnote{\url{https://github.com/hzwer/NIPS2017-LearningToRun}}.
  
We built the ensemble by drawing models trained with settings of the last section. Fig.~\ref{fig:ace}(b) gives the distribution of rewards when using ACE, where AXCY stands for X number of actors and Y number of critics. It can be seen that A10C10 (having 10 critics and 10 actors) has a much smaller chance of falling (rewards below 30) compared to A1C0, which is equivalent to DDPG. The maximum rewards also get improved, as shown in Tab.~\ref{tab:ace}.

Training with ACE was found to have similar performance as Inference-Time ACE.

\begin{table}
\centering
\caption{Performance of ACE}
\label{tab:ace}
\begin{tabular}{@{}ccccclc@{}}
\toprule
Experiment & \# Test & \# Actor & \# Critic & Average reward & Max reward & \# Fall off \\ \midrule
A1C0 & 100 & 1 & 0 & 32.0789 & 41.4203 & 25 \\
A10C1 & 100 & 10 & 1 & 37.7578 & 41.4445 & 7 \\
A10C10 & 100 & 10 & 10 & 39.2579 & 41.9507 & 4 \\ \bottomrule
\end{tabular}
\end{table}

\subsection{Discussion}
We propose Actor-Critic Ensemble, a deterministic method that avoids dooming actions at inference time by asking an ensemble of critics to pick actions proposed by an ensemble of actors. Our experiments found that ACE can significantly improve performance of DDPG, by reducing of the number of falls and increasing the speed of running skeletons.

\section{Deep Deterministic Policy Gradient and improvements}\label{s:reason8}
\sectionauthor{Mikhail Pavlov, Sergey Kolesnikov and Sergey Plis}

We benchmarked state of the art policy-gradient methods and found that Deep Deterministic Policy Gradient (DDPG) method is the most efficient method for this environment. We also applied several improvements to DDPG method, such as layer normalization, parameter noise, action and state reflecting. All this improvements helped to stabilize training and improve its sample-efficiency. 
\subsection{Methods}
\subsubsection{DDPG improvements}
We used standard reinforcement learning techniques: action repeat (the agent selects action every 5th state and selected action is repeated on skipped steps) and reward scaling. After several attempts, we choose a scale factor of 10 (i.e. multiply reward by ten) for remaining experiments.
For exploration we used Ornstein-Uhlenbeck (OU) process~\cite{uhlenbeck1930theory} to generate temporally correlated noise, considered efficient in exploration of physical environments.
Our DDPG implementation was parallelized as follows: $n$ processes collected samples with fixed weights all of which were processed by the learning process at the end of an episode, which updated their weights.
Since DDPG is an off-policy method, the stale weights of the samples only improved the performance providing each sampling process with its own weights and thus improving exploration.

\subsubsection{Parameter noise}
Another improvement is the recently proposed parameters
noise~\cite{plappert2017parameter} that perturbs network weights
encouraging state dependent exploration.
We used parameter noise only for the actor network.
Standard deviation $\sigma$ for the Gaussian noise was chosen according
to the original work~\cite{plappert2017parameter} so that measure
where $\widetilde{\pi}$ is the policy with noise, equals to $\sigma$ in OU.
For each training episode we switched between the action noise and the parameter noise choosing them with $0.7$ and $0.3$ probability respectively.

\subsubsection{Layer norm}
Henderson et al. showed that layer normalization~\cite{ba2016layer} stabilizes the learning process in a wide range of reward scaling.
We have investigated this claim in our settings.
Additionally, layer normalization allowed us to use same perturbation scale across all layers despite the use of parameters noise~\cite{plappert2017parameter}.
We normalized the output of each layer except the last for critic and actor by standardizing the activations of each sample.
We applied layer normalization before the nonlinearity.

\subsubsection{Actions and states reflection symmetry}
The musculoskeletal model to control in the challenge has bilateral body symmetry. State components and actions
can be reflected to increase sample size by factor of 2.  We sampled
transitions from replay memory, reflected states and actions and used
original states and actions as well as reflected  as batch in training
step. This procedure improves stability of learned policy. When we did not
use this technique our model learned suboptimal policies, when for example
muscles for only one leg are active and other leg just follows the first
leg.

\subsection{Experiments and results}
For all experiments we used environment with 3 obstacles and random strengths of the psoas muscles.
We tested models on setups running 8 and 20 threads.
For comparing different PPO, TRPO and DDPG settings we used 20 threads per model configuration.
We have compared various combinations of improvements of DDPG in two identical settings that only differed in the number of threads used per configuration: 8 and 20.
The goal was to determine whether the model rankings are consistent when the number of threads changes.
For $n$ threads (where $n$ is either 8 or 20) we used $n-2$ threads for sampling transitions, 1 thread for training, and 1 thread for testing.
For all models we used identical architecture of actor and critic networks.
All hyperparameters are listed in Table~\ref{tab:hyperopt8}.
Our code used for competition and described experiments can be found
in a github repo.\footnote{Theano:
  \url{https://github.com/fgvbrt/nips_rl} and  PyTorch: \url{https://github.com/Scitator/Run-Skeleton-Run}} 
Experimental evaluation is based on the non-discounted return.

\begin{table}
\caption{Hyperparameters used in the experiments.}
\begin{tabularx}{\columnwidth}{r|X}
  \toprule
  parameters & Value \\
  \midrule
  Actor network architecture &  $[64,64]$, elu activation\\
  Critic network architecture & $[64,32]$, tanh activation \\
  Actor learning rate & linear decay from $1\mathrm{e}{-3}$ to
                        $5\mathrm{e}{-5}$ in $10\mathrm{e}{6}$ steps with Adam optimizer\\
  Critic learning rate & linear decay from $2\mathrm{e}{-3}$ to
                        $5\mathrm{e}{-5}$ in $10\mathrm{e}{6}$ steps with Adam optimizer\\
  Batch size & 200\\
  $\gamma$ & 0.9 \\
  replay buffer size & $5\mathrm{e}{6}$\\
  rewards scaling & 10 \\
  parameter noise probability & 0.3 \\
  OU exploration parameters & $\theta = 0.1$, $\mu = 0$, $\sigma = 0.2$,
                          $\sigma_{min}=0.05$, $dt = 1\mathrm{e}{-2}$,
                          $n_\text{steps}$ annealing
                          $\sigma_\text{decay} 1\mathrm{e}{6}$ per thread\\
                          \bottomrule
\end{tabularx}
\label{tab:hyperopt8}
\end{table}

\subsubsection{Benchmarking different models}
\begin{figure}[t]
\centering
\includegraphics[width=0.6\columnwidth]{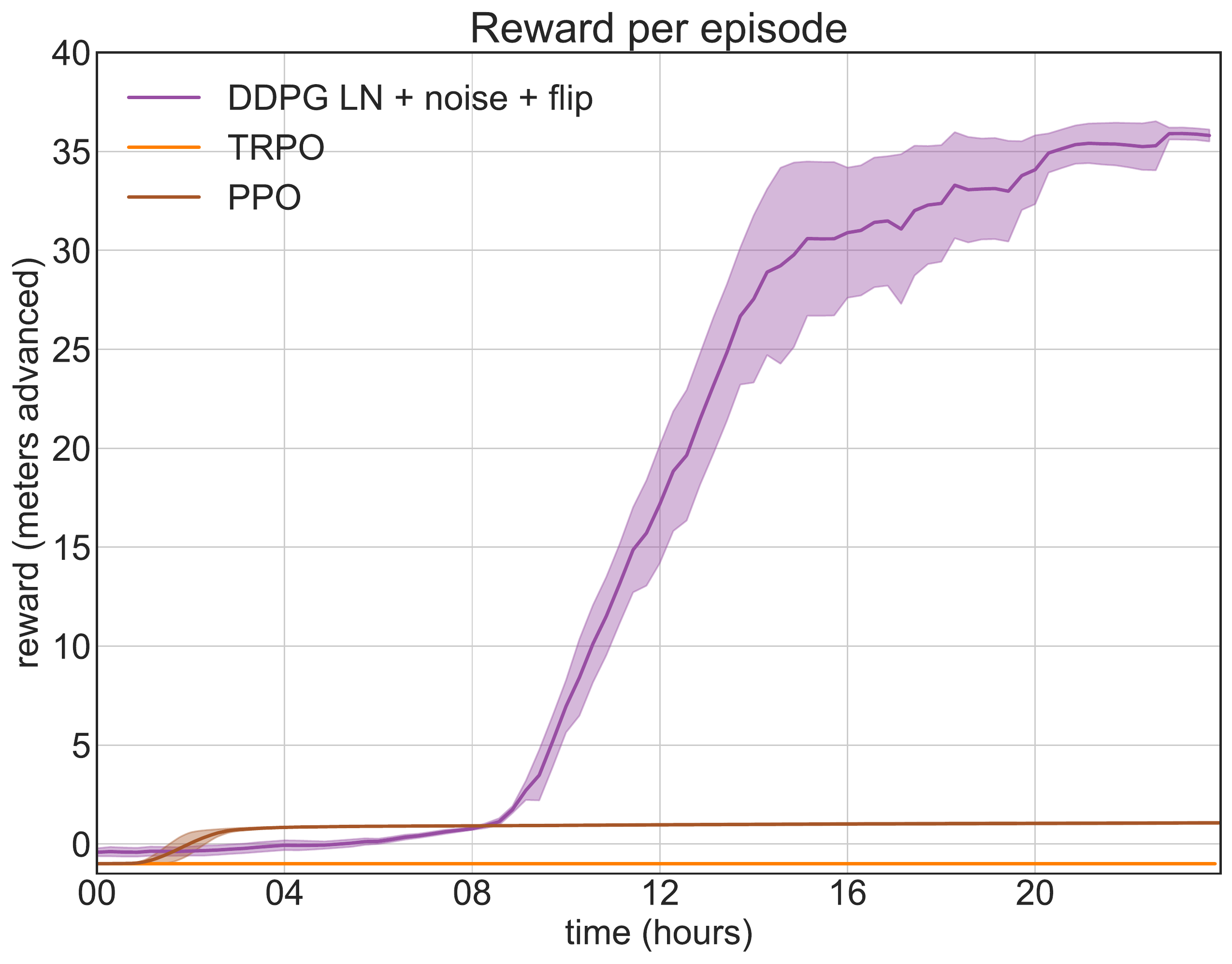}
\caption{Comparing test reward of the baseline models and the best performing model that we have used in the ``Learning to run'' competition.}
\label{fig:baseline_compare}
\end{figure}
Comparison of our winning model with the baseline approaches is presented in Figure~\ref{fig:baseline_compare}.
Among all methods the DDPG significantly outperformed PPO and TRPO.
The environment is time expensive and method should utilized experience as effectively as possible.
DDPG due to experience replay (re)uses each sample from environment many times making it the most effective method for this environment.

\subsubsection{Testing improvements of DDPG}

\begin{figure}
    \begin{subfigure}[t]{0.5\textwidth}
	\centering
      \includegraphics[width=0.95\textwidth]{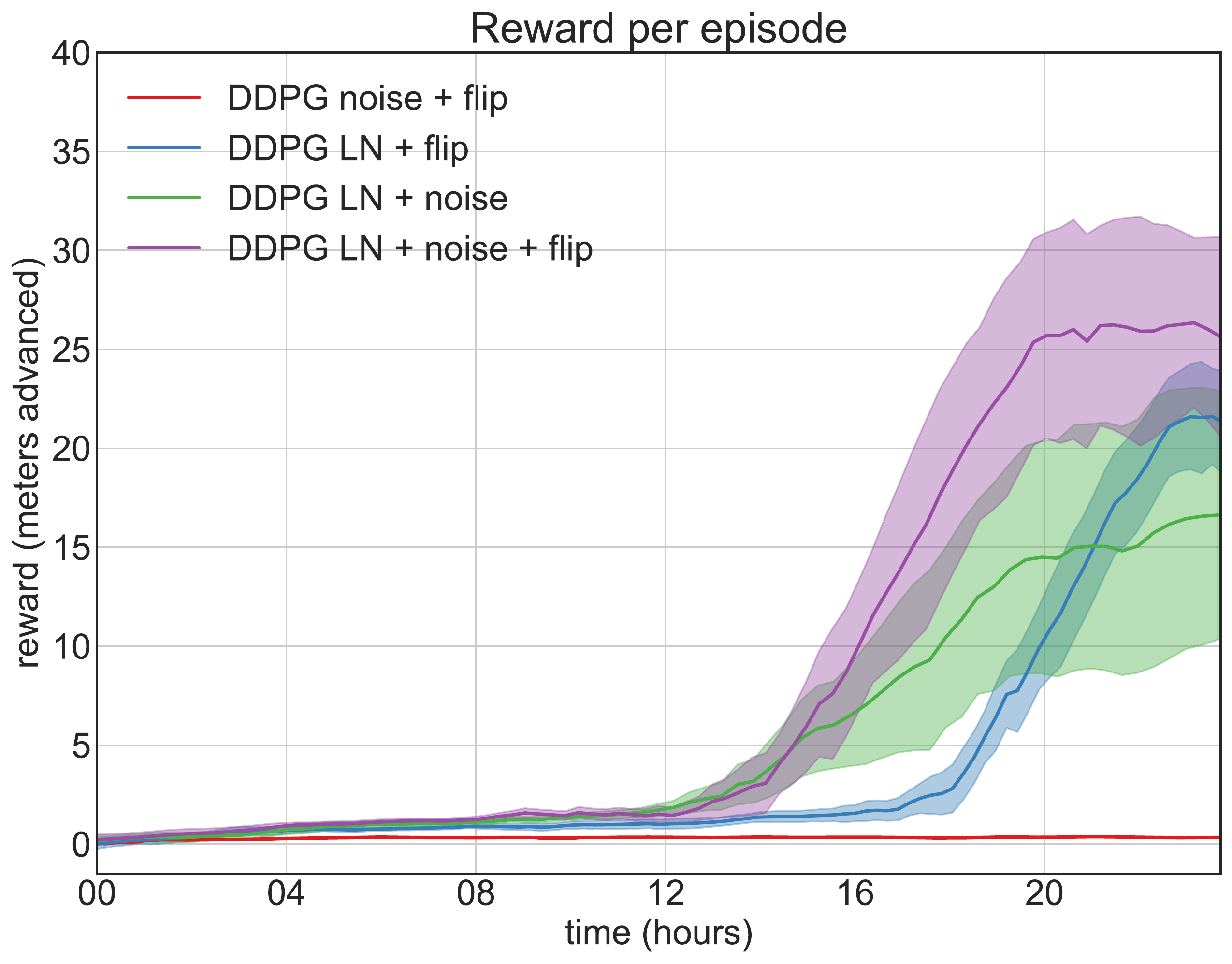}
      \caption{$8$ threads}
            \label{fig:8_threads}
    \end{subfigure}
    \begin{subfigure}[t]{.5\textwidth}
    \centering
	\includegraphics[width=0.95\textwidth]{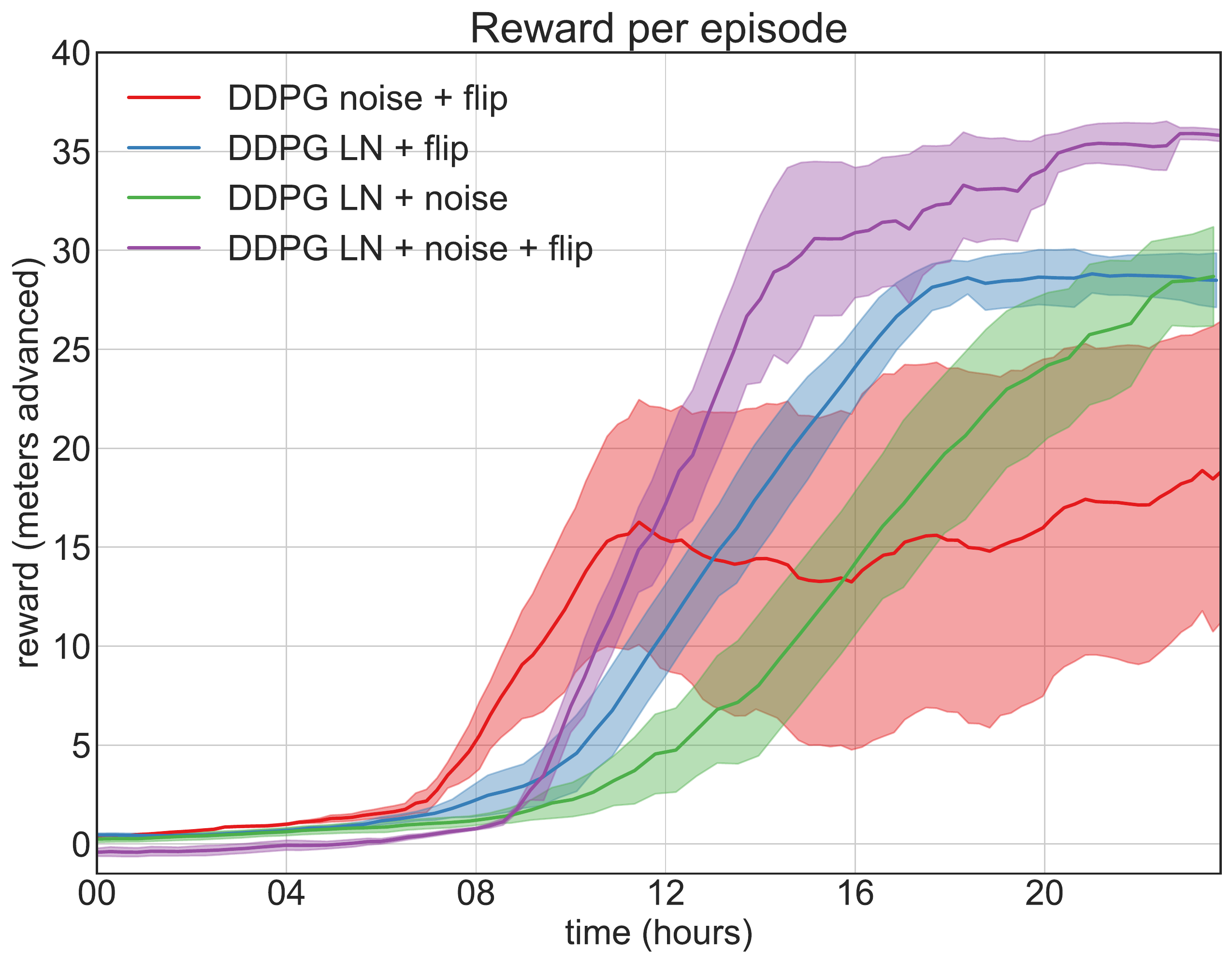}
      \caption{$20$ threads}
      \label{fig:20_threads}
    \end{subfigure}
    
    \caption{Comparing test reward for various modifications of the DDPG algorithm with 8 threads per configuration (Figure~\ref{fig:8_threads}) and 20 threads per configuration (Figure~\ref{fig:20_threads}). Although the number of threads significantly affects performance, the model ranking approximately stays the same.}
    \label{fig:ablation}
\end{figure}

To evaluate each component we used an ablation study as it was done in the rainbow article~\cite{hessel2017rainbow}.
In each ablation, we removed one component from the full combination.
Results of experiments are presented in Figure~\ref{fig:8_threads} and Figure~\ref{fig:20_threads} for 8 and 20 threads respectively.
The figures demonstrate that each modification leads to a statistically significant performance increase.
The model containing all modifications scores the highest reward.
Note, the substantially lower reward in the case, when parameter noise was employed without the layer norm.
One of the reasons is our use of the same perturbation scale across all layers, which does not work that well without normalization.
Also note, the behavior is quite stable across number of threads, as
well as the model ranking.
As expected, increasing the number of threads improves the result.

Maximal rewards achieved in the given time for 8 and 20 threads cases for each of the combinations of the modifications is summarized in Table~\ref{tab:maxreward}.
The main things to observe is a substantial improvement effect of the number of threads, and stability in the best and worst model rankings, although the models in the middle are ready to trade places.

\begin{table}
\caption{Best achieved reward for each DDPG modification.}
\begin{tabularx}{\columnwidth}{c|X|X}
  \toprule
  \backslashbox{agent}{\# threads} & 8 & 20 \\
  \midrule
  DDPG + noise + flip & 0.39 & 23.58 \\
  DDPG + LN + flip & 25.29 & 31.91\\
  DDPG + LN + noise & 25.57 & 30.90 \\
  DDPG + LN + noise + flip & \bf{31.25} & \bf{38.46}\\
  \bottomrule
\end{tabularx}
\label{tab:maxreward}
\end{table}

\subsection{Discussion}
Our results in OpenSim experiments indicate that in a computationally expensive stochastic environments that have high-dimensional continuous action space the best performing method is off-policy DDPG.
We have tested 3 modifications to DDPG and each turned out to be important for learning.
Action states reflection doubles the size of the training data and improves stability of learning and encourages the agent to learn to use left and right muscles equally well.
With this approach the agent truly learns to run.
Examples of the learned policies with and without the reflection are present at this URL \url{https://tinyurl.com/ycvfq8cv}.
Parameter and Layer noise additionally improves stability of learning due to introduction of state dependent exploration.

\section{Asynchronous DDPG with Deep Residual Network for Learning to Run}\label{s:imcl}\sectionauthor{Zhibo Chen, Zhizheng Zhang, Jiale Chen and Jun Shi}

For improving the training effectiveness of DDPG on this physics-based simulation environment which has high computational complexity, we designed a parallel architecture with deep residual network for the asynchronous training of DDPG. In this work, we describe our approach and we introduce supporting implementation details.

\subsection{Methods}
\subsubsection{Asynchronous DDPG}
In our framework, the agent could collect interactive experiences and update its network parameters asynchronously. For the collection of experiences, the \textit{Learning to Run} environments with different seeds and same difficulty-level settings were wrapped by multi-process programming. All step-by-step interactive experiences in every wrapped environment would be stored in a specific storage until this episode finished. Then we decided which step experiences to put into the experience replay memory according to their corresponding step rewards and episode rewards. In terms of the updating of networks' parameters, the updating process would sample a batch from the replay memory after each interaction with the RL environments no matter which specific environment process this interaction takes place in.

\subsubsection{The Neural Network Structure}
\begin{figure}[h]
	\centering
	\includegraphics[scale=0.4]{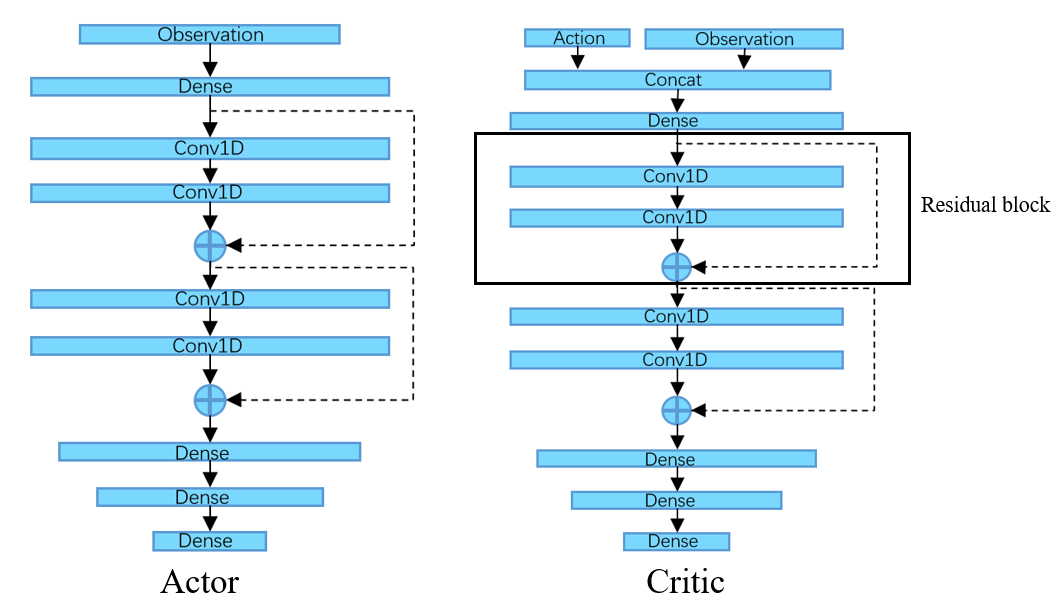}
	\caption{The diagram of our network structure with residual blocks}
	\label{fig-network}
\end{figure}

Whether for the human body in real-world or the musculoskeletal model used in this simulation, the accurate physical motions are determined by multiple joints and implemented by the excitations of multiple different muscles. Taking it naturally, we applied 1D convolution modules in the neural networks for both actor and critic networks with the expectation of capturing the correlation among 41 values of the observation. And our experimental results indicated that 1D convolution neural networks were better able to prevent converging to the local optimal solution than fully connected networks. In order to improve the efficiency and stability of training, we added the residual blocks (see Figure ~\ref{fig-network}) to make our model easier to train and converge. \\

We also tried to take advantage of 2D convolution to process the 1D observation information and learn the features from historical actions, inspired by the work on 3D convolution neural network for human action recognition ~\cite{ji20133d}. However, the performance of the RL agent with 2D convolution was less likely to converge steadily.

\subsubsection{Noise for Exploration}
We tried both parameter space noise~\cite{plappert2017parameter} and action space noise for the exploration in this task. For parameter space noise, it was really hard to fine-tuning and get an optimal solution in \textit{Learning to Run} environment, which might be caused by the structure of our neural network. In terms of action space noise, we found that Ornstein-Uhlenbeck noise would lead to inefficient exploration and convergence to local optimal solution. Instead, correlated Gaussian noise was more suitable in this task. Additionally, considering there should be the continuity among actions as the outputs of the actor, we designed a so-called random walk Gaussian noise for this continuous task, which brought us the highest grade but with a little large variance. Hence, we thought that normal correlated Gaussian noise and the random walk Gaussian noise were both effective for exploration in this environment, but each had its advantages and disadvantages.

\subsubsection{Detailed Implementation}
We would like to discuss the stability in this section, which included the stability of training and the stability of the policy for obstacles crossing. For the stability of training, we applied some common techniques in our model, such as layer normalization, reward scaling, prioritized experience replay~\cite{schaul2015prioritized} and action repetition. Additionally, we applied a training trick with a small learning rate, named ``trot''. In detail, we sampled one batch and used it to implement back-propagation for multiple times with a small learning rate. For the stability of the policy for obstacles crossing, a ceil option for the radius of the obstacle turned out to be significant to improve the agent's performance. The obstacles for the agent would be slightly larger than their real sizes and the mathematic space to be fitted by neural network will be also reduced by this method. Hence, we could make it easier for the learning by neural network and improve the stability of obstacles crossing meanwhile.

\subsection{Experiments and results}
\subsubsection{The Number of Parallel Process}
Based on the experimental verification, we found the number of environment process would have a large impact on the learning performance of the agent. We tested our model with setting 12, 24, and 64 processes. The experimental results indicated that the more processes we used, the more samples we could get but not inevitably the better for the learning performance. In our settings, the model with 24 environment process would get the highest grade. Excessive number of the parallel environment processes could cause that too much similar transitions were pushed into the experience replay memory, which adversely affected the training of agent.

\subsubsection{The Neural Network Structure}
\begin{figure}[h]
	\centering
	\includegraphics[scale=0.3]{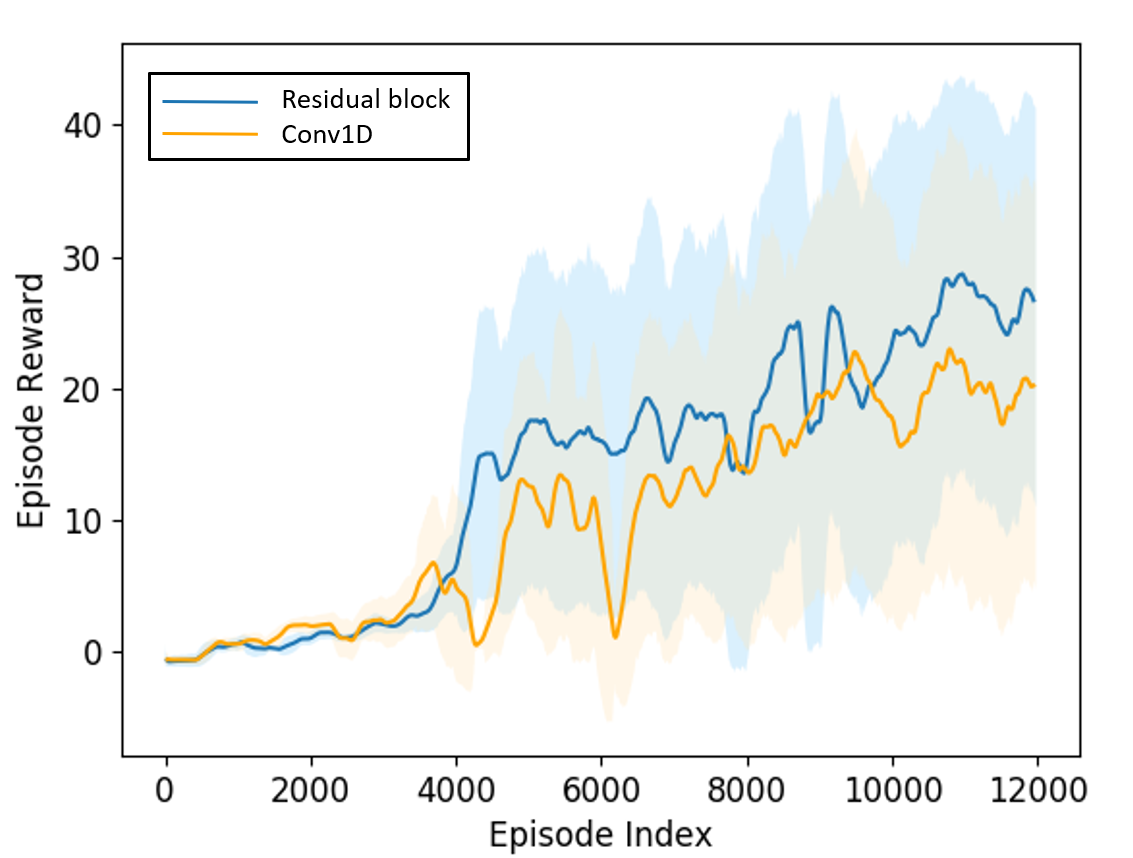}
	\caption{The results of the comparison of network structure with and without residual module}
	\label{fig-result}
\end{figure}

The neural network structure seriously affects the learning ability of the agent. According to our training results (see Figure ~\ref{fig-result}), the 1D convolution neural network with residual blocks was the best fit for both actor and critic in DDPG whether for the maximum learning ability or the stability. Moreover, the 1D convolution neural network without residual blocks was also better than the fully connection network. Moreover, widening the network was more effective than deepening the network in this task, because it was easier to get the gradient in a suitable range for wide networks to implement policy updating.

\subsection{Discussion}
In this section, we discuss the difference between the first and the second round of the competition. Due to the setting of that the current episode will be finished if you fall down, the requirements on stability of obstacles crossing were significantly higher than it in the first round. More precisely, the obstacles only appeared in the beginning of the journey, it was easier for an agent to cross the obstacles in the set time, because the agent had not accelerated yet to a great speed. In the second round, the agent should cross the obstacles when it runs with a high speed, which made it easy to fall down. As our previous description, a ceiling constraint for the radius of obstacles could solve this problem effectively to some extent.

Interestingly, our agent discovered a really smart trick by itself in the training process. Instead of avoiding the obstacles, it would try to deliberately step on a large obstacle and then used it as a stepping stone. Although it didn't master this trick due to the limitation of training time, we were really surprised by this exploitation of the environment.

\section{Proximal Policy Optimization with Policy Blending}\label{s:deepsense}
\sectionauthor{Henryk Michalewski, Piotr Miłoś and Błażej Osiński}

Our solution was based on the distributed Proximal Policy Optimization algorithm combined with a few efficiency-improving techniques. We used the \textit{frameskip} to increase exploration. We changed rewards to encourage the agent to \textit{bend its knees}, which significantly stabilized the gait and accelerated the training. In the final stage, we found it beneficial to transfer skills from  small networks (easier to train) to bigger ones (with more expressive power). For this purpose we developed \textit{policy blending}, a general cloning/transferring technique.

\subsection{Methods}

\paragraph*{Combining cautious and aggressive strategies.}

In the first stage of the competition our most successful submission was a combination of 2 agents. The first agent was cautious and rather slow. It was designed to steadily jump over the three obstacles (approx. first 200 steps of an episode). For the remaining 800 steps we switched to a 20\%-faster agent, trained beforehand in an   environment without obstacles.

\begin{figure}[H]
\begin{tikzpicture}[scale=0.4]
\draw (0,0) -- (33,0);

\foreach \x in {0.8,5,7.5,10,32.2}
\draw(\x cm,3pt) -- (\x cm, -3pt);

\node[align=center] at (5.4,1.5) {\textbf{\small jumper policy}};
\draw [thick,decorate,decoration={brace,amplitude=6pt,raise=0pt}] (0.8,0.3) -- (10,0.3);

\node[align=center] at (12.5,1.5) {\textbf{\small mixed policy}};
\draw [thick,decorate,decoration={brace,amplitude=6pt,raise=0pt}] (10,0.3) -- (15,0.3);

\node[align=center] at (23.6,1.5) {\textbf{\small sprinter policy}};
\draw [thick,decorate,decoration={brace,amplitude=6pt,raise=0pt}] (15,0.3) -- (32.2,0.3);

\draw (0.8,0) node[below=3pt] {\textbf{\small start}};
\draw (5,0) node[below=3pt] {$O_1$};
\draw (7.5,0) node[below=3pt] {$O_2$};
\draw (10,0) node[below=3pt] {$O_3$};
\draw (32.2,0) node[below=3pt] {\textbf{\small end}};
\end{tikzpicture}
\end{figure}

The switching of policies was a rather delicate task. Our attempts to immediately change from the first policy left the agent in a state unknown to the second one and caused the agent to fall. Eventually, we switched the policies gradually by applying the linear combination $(1-\frac{k}{n})a_\eta + \frac{k}{n}a_\nu$, where $k$ is the transition step, $a_\eta, a_\nu$ are actions of the jumper and sprinter respectively; $n=150$ was required to smooth the transition. 
A more refined version of this experiment should include learning of a macro policy which would decide on its own how the jumper and sprinter should be combined (see \cite{Sutton1999,hq} for a broader introduction to hierarchical learning).

\begin{minipage}{0.45\textwidth}
\begin{tikzpicture}
\node [anchor=west] (note) at (2,7.7) {\textbf{{\large angle $\alpha$}}};
\begin{scope}[xshift=0.5cm]
    \node[anchor=south west,inner sep=0] (image) at (0,0) {\includegraphics[width=5.5cm]{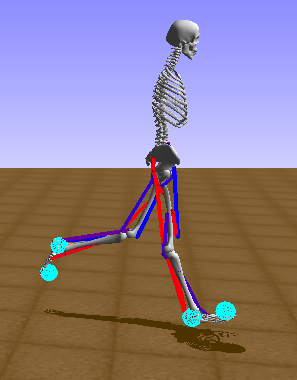}};
    \begin{scope}[x={(image.south east)},y={(image.north west)}]
        \draw [-latex, ultra thick, yellow] (note) to[out=180, in=120] (0.43,0.4);
    \end{scope}
\end{scope}
\end{tikzpicture}
\end{minipage}
\hfill
\begin{minipage}{0.5\textwidth}
\paragraph*{Frameskip.}
Our initial experiments led to suboptimal behaviors, such as kangaroo-like jumping (two legs moving together). We conjectured that the reason was a poor exploration and thus applied frameskip. In this way we obtained our first truly bipedal walkers. In particular, frameskip set to $4$ led to a slow but steadily walking agent capable of scoring approximately 20 points. 

\paragraph*{Reward shaping.} 
In the process of training we obtained various agents walking or running, but still never bending one of the legs. We decided to {\em shape agent's behavior} \cite{robot_shaping} through adding of an extra reward for bending of knees.
More specifically, we added a reward for getting the angle $\alpha$ into a prescribed interval, see the figure on the left.
This adjustment resulted in significant improvements in the walking style.
Moreover, training with the knee reward caused our policy to train significantly faster, see Figure \ref{fig:knee}.
We experimented with various other rewards. Perhaps the most interesting was explicit penalization of falling over. As a consequence we created an ultra-cautious agent, who always concluded the whole episode but at the cost of a big drop of the average speed.
\end{minipage}

\paragraph*{Final tuning - policy blending.}
We found it easier to train a reasonable bipedal walking policy~$\pi$ when using small nets. However, small nets suffered from unsatisfactory final performance in environment with many obstacles. We used pretrained $\pi$ and a method we called \emph{policy blending} to softly guide the learning process. Namely, we kept $\pi$ fixed and trained new $\eta$; the agent was issuing actions $\alpha a_{\pi} + (1 - \alpha) a_{\eta}$, $\alpha\in(0,1)$. One can see blending as  a simplified version of progressive networks considered in \cite{progressive}.

Even with $\alpha\approx 0.1$ the walking style of $\pi$ was coarsely preserved, while the input from bigger net of $\eta$ led to significant improvements in harder environments.
In some cases blended policies could successfully deal with obstacles even though $\pi$ was trained in an obstacle-free environment.
In some experiments after an initial period of training with $\alpha>0$, we continued with $\alpha=0$, which can be seen as knowledge transfer from $\pi$ to $\eta$.
In our experience, such knowledge transfer worked better than direct behavioral cloning.
\begin{wrapfigure}{r}{0.43\textwidth}
\begin{tikzpicture}[every text node part/.style={align=center}, scale=0.41]
	  {\small
      \node(pi) [draw,rectangle] at (0,0) {$\pi$ \\ $\text{hid}\_\text{size} = 32$};
      \node(eta) [draw,rectangle] at (10,0) {$\eta$ \\ $\text{hid}\_\text{size} = 128$};
      \node(sum) [rectangle] at (5,5) {$\alpha a_{\pi} + (1 - \alpha) a_{\eta}$};
      }
      \draw[->] (pi) to (sum);
      \draw[->] (eta) to (sum);
\end{tikzpicture}
\captionof{figure}{Blending of $\pi$ and $\eta$.}
\end{wrapfigure}
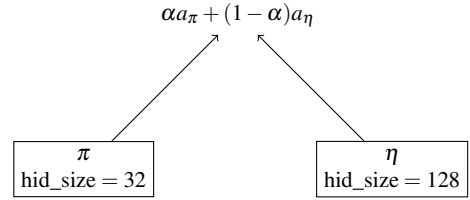

To test cloning we followed a procedure of \cite{cloning}, first copying the original policy to a new neural net through supervised learning followed by further training of the new policy on its own.
However, in our experiments    policies obtained through cloning  showed at best the same performance as the original policies.
Conversely, as can be seen in Figure \ref{fig:policy} transferring a policy from obstacle-free environment using policy blending performed better than simple retraining.

\subsection{Experiments and results}

\begin{minipage}{0.45\textwidth}
\includegraphics[width=\linewidth]{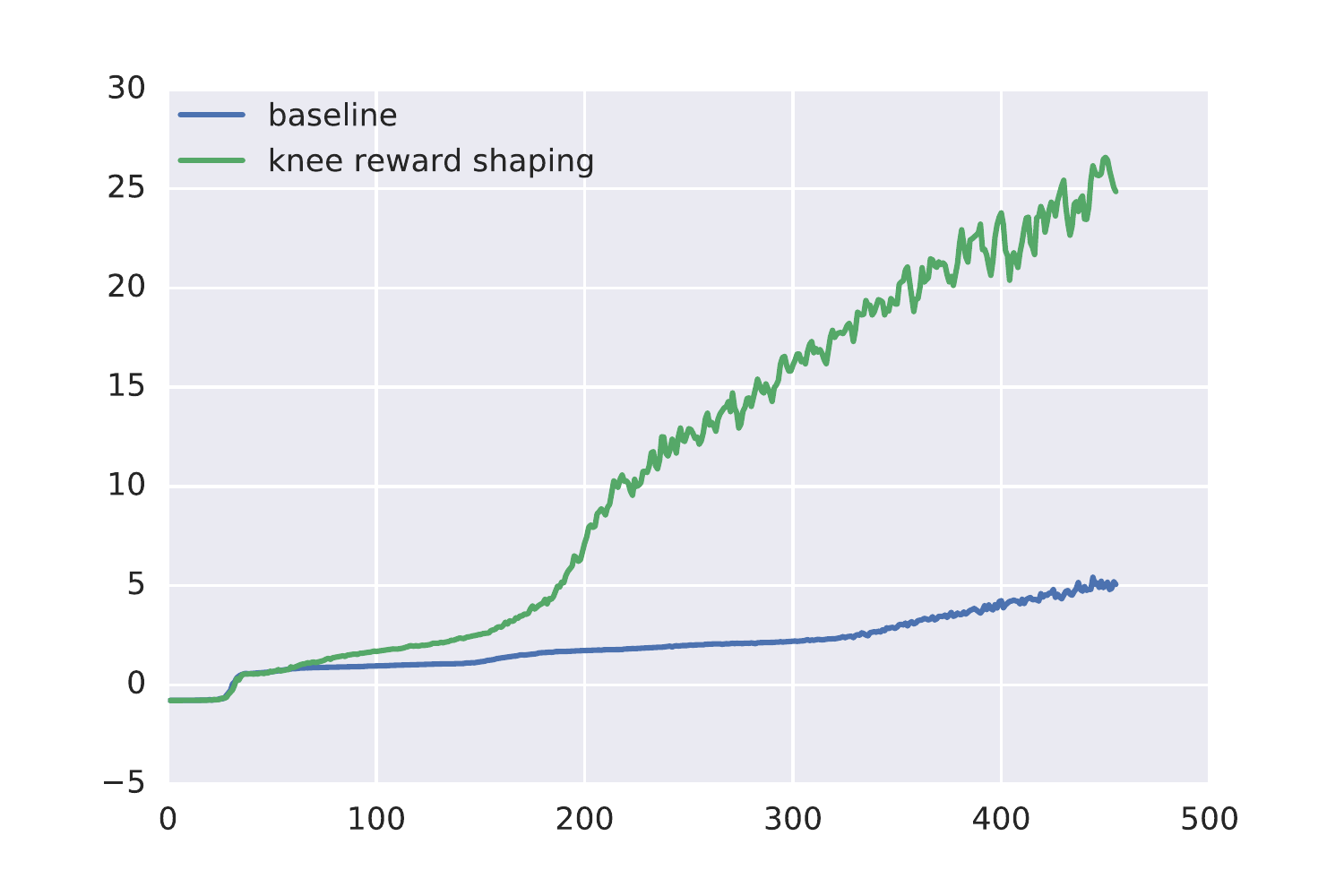}
\captionof{figure}{Reward shaping}\label{fig:knee}
\end{minipage}\hfill
\begin{minipage}{0.45\textwidth}
	\includegraphics[width=\linewidth]{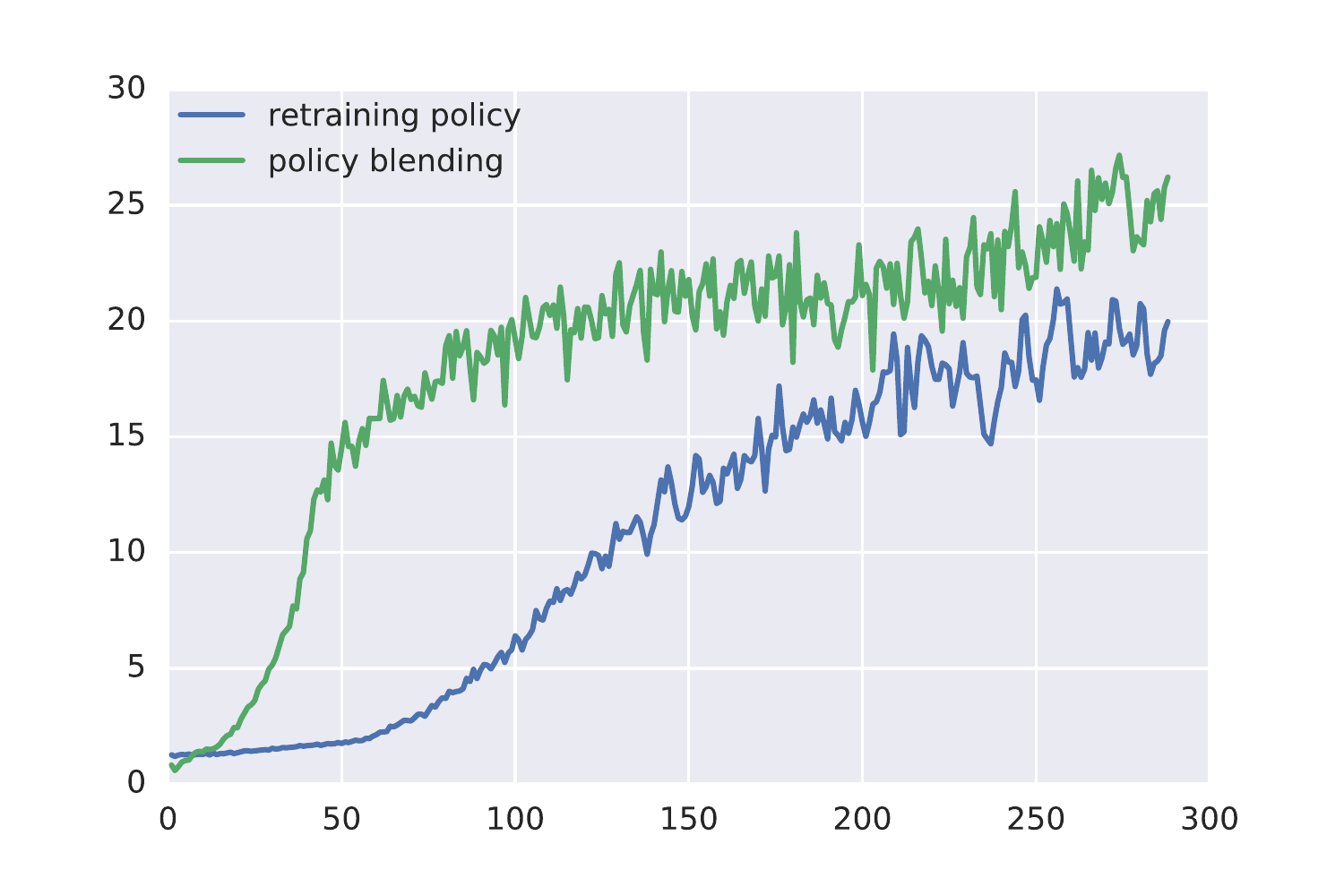}
\captionof{figure}{Policy blending}\label{fig:policy}
\end{minipage}
\medskip

Overall, we performed approximately $5000$ experiments of lengths up to 72 hours. We used the PPO optimization algorithm \cite{ppo}. In Figure \ref{fig:knee} we present how the knee reward helps in the training. Figure \ref{fig:policy} we compare retraining with blending of policies.

\section{Double Bootstrapped DDPG for Continuous Deep Reinforcement Learning}\label{s:jgeek}
\sectionauthor{Zhuobin Zheng, Chun Yuan and Zhihui Lin}

Deep Deterministic Policy Gradient (DDPG) provides substantial gains in sample efficiency on off-policy experience data over on-policy methods. However, vanilla DDPG may explore inefficiently and be easily trapped into local optima in the Learning to Run Challenge. We proposed \textit{Double Bootstrapped DDPG}, an algorithm that combines efficient exploration with promising stability in continuous control tasks.
\subsection{Methods}
\subsubsection{Double Bootstrapped DDPG}
\begin{figure}[htbp]
	\centering
	\includegraphics[scale=0.8]{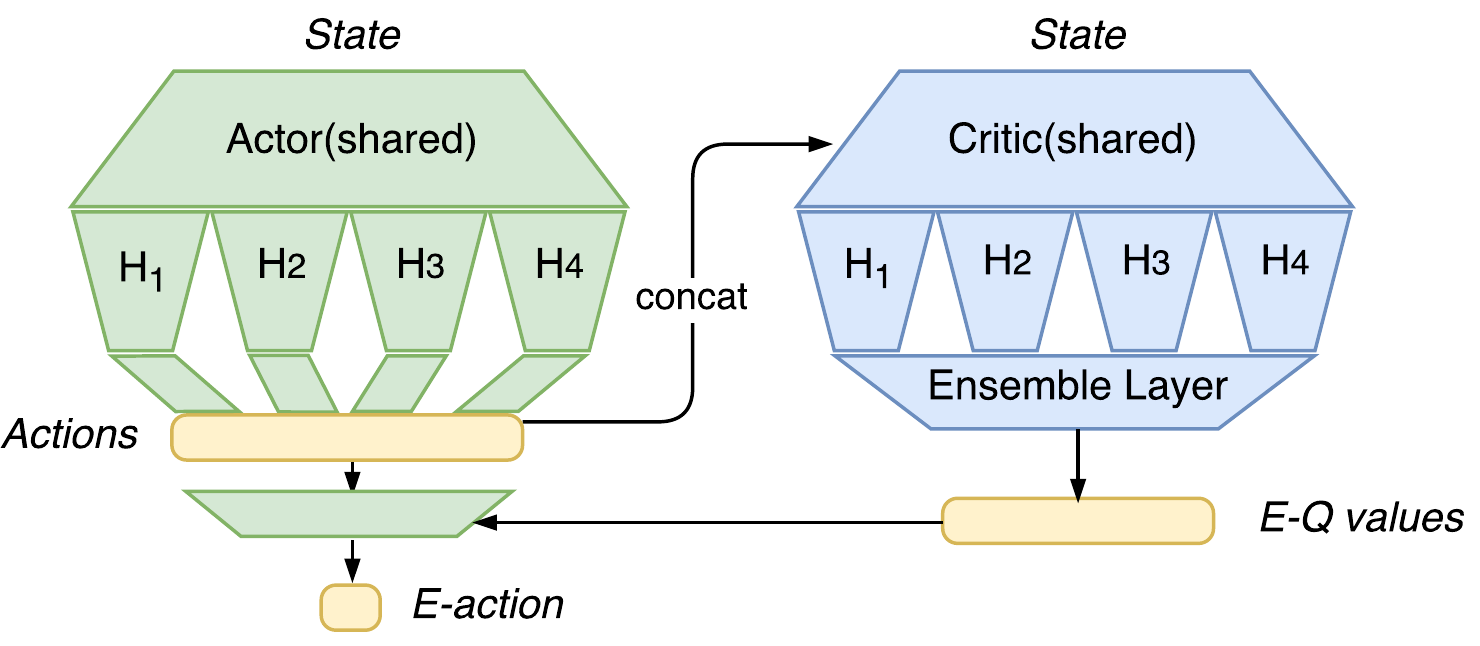}
	\caption{Structure of Double Bootstrapped DDPG. When the actor(green) receives a state, each head outputs an action vector($\mathbb{R}^{18}$). Given the same state, the critic(blue) concatenates these actions($\mathbb{R}^{K{\times}18}$) in the hidden layer and outputs an $Ensemble$-$Q$ value matrix($\mathbb{R}^{K{\times}K}$) by $K$ heads. The actor chooses the final $Ensemble$-action determined by the $Ensemble$-$Q$ values.}\label{fig:fig-dbddpg}
\end{figure}

Inspired by Bootstrapped DQN~\cite{osband2016deep}, Double Bootstrapped DDPG, abbreviated with DB-DDPG, extends the actor-critic architecture to completely bootstrapped networks (see Figure~\ref{fig:fig-dbddpg} for an overview of the approach). Both actor and critic networks have a shared body for feature extraction, followed by multiple heads with different random initialization.

A simple warm-up is applied before training: the actor heads are randomly selected to interact with the environment and pre-train during every episode, together with their paired critic heads as vanilla DDPGs.

During one single training episode, the $k_{th}$ pair of actor and critic heads are randomly activated to learn. Given a state $s_t$, multiple actor heads output candidate actions $\textbf{\textit{a}}_t=(a^1,a^2,...,a^K)_{t}$, which are concatenated to the critic network. Multiple critic heads output an $E$-$Q$ value matrix($\mathbb{R}^{K{\times}K}$) for the actions $\textbf{\textit{a}}_t$ according to the state $s_t$. The final ensemble action with highest $Q$-value sum which is determined by the $E$-$Q$ value matrix as Equation~\eqref{eq:action}, 
\begin{equation}\label{eq:action}
a_t=\argmax_a\{\sum_{i=1}^{K}Q_{i}(s_t,a)|_{{a=\mu_k(s_t)}}\}_{k=1}^K,
\end{equation}
is chosen to execute for the state receiving a new state $s_{t+1}$ and a reward $r_t$.

Moreover, a random mask $\textbf{\textit{m}}_t=(m^1,m^2,...,m^K)_t$ is generated with Bernoulli distribution simultaneously. A new transition $(s_t,a_t,r_t,s_{t+1},\textbf{\textit{m}}_t)$ is stored in experience memory. The selected $k_{th}$ heads together with respective shared bodies and target networks are trained as a DDPG given a minibatch of samples. The $i_{th}$ experience with mask $m_{i}^k=0$ is ignored when training for bootstrapping~\cite{osband2016deep}. A more detailed procedure can be found in Algorithm~\ref{code-dbddpg}. 

Besides, we also used common reinforcement learning techniques, such as: frame skipping~\cite{DBLP:journals/corr/MnihKSGAWR13}(the agent performs the same action every $k$ consecutive observations. we set $k=4$), prioritized experience replay~\cite{schaul2015prioritized} and a reward trick~\cite{heess2017emergence} which utilizes velocity instead of distance as reward encouraging the agent to make more forward process along the track. 
	
\subsubsection{Observation Preprocessing}
Since the original observation vector given by the environment does not satisfy the Markov property, we extended the observation by calculating more velocities and acceleration of the remaining body parts. During the experiments, we found that obstacles may be extremely close to each other. In this case, when an agent overstepped the first obstacle with one single leg, a new observation about the next obstacle was immediately updated. As a result, it probably fell down since another leg hit the previously invisible obstacle. To solve this problem, information about the previous obstacle was appended to the observation vector.

\subsubsection{Noise Schedule}
Ornstein-Uhlenbeck (OU) process~\cite{uhlenbeck1930theory} was used to generate noise for exploration. DB-DDPG utilized a noise decay rate for action noise for balancing exploration and exploitation. Once the noise decreased below a threshold, DB-DDPG acted without noise injection and focused on exploiting the environment. Exploration noise recovered when the uncertainty of the agent decreased. This process could be switched iteratively and fine-tuned manually for stability until convergence.

\subsection{Experiments}
\subsubsection{Details}
We used Adam~\cite{DBLP:journals/corr/KingmaB14} for learning the parameters with a learning rate of $1e^{-4}$ and $3e^{-4}$ for the actor and the critic network respectively. For the $Q$ networks we set a discount factor of $\gamma=0.99$ and $\tau=1e^{-3}$ for soft target updates. All hidden layers utilized exponential linear units(ELU)~\cite{clevert2015fast} while the output layer of the actor utilized a $tanh$ layer to bound the actions followed by scale and shift operations. The shared network of the actor had two hidden layers with 128 and 64 units respectively. This was followed by multiple heads with 64 and 32 units. The critic had similar architecture except actions which were concatenated in the second hidden layer. 

\subsubsection{Results}
\begin{figure}[hbtp]
	\centering
	\includegraphics[scale=0.6]{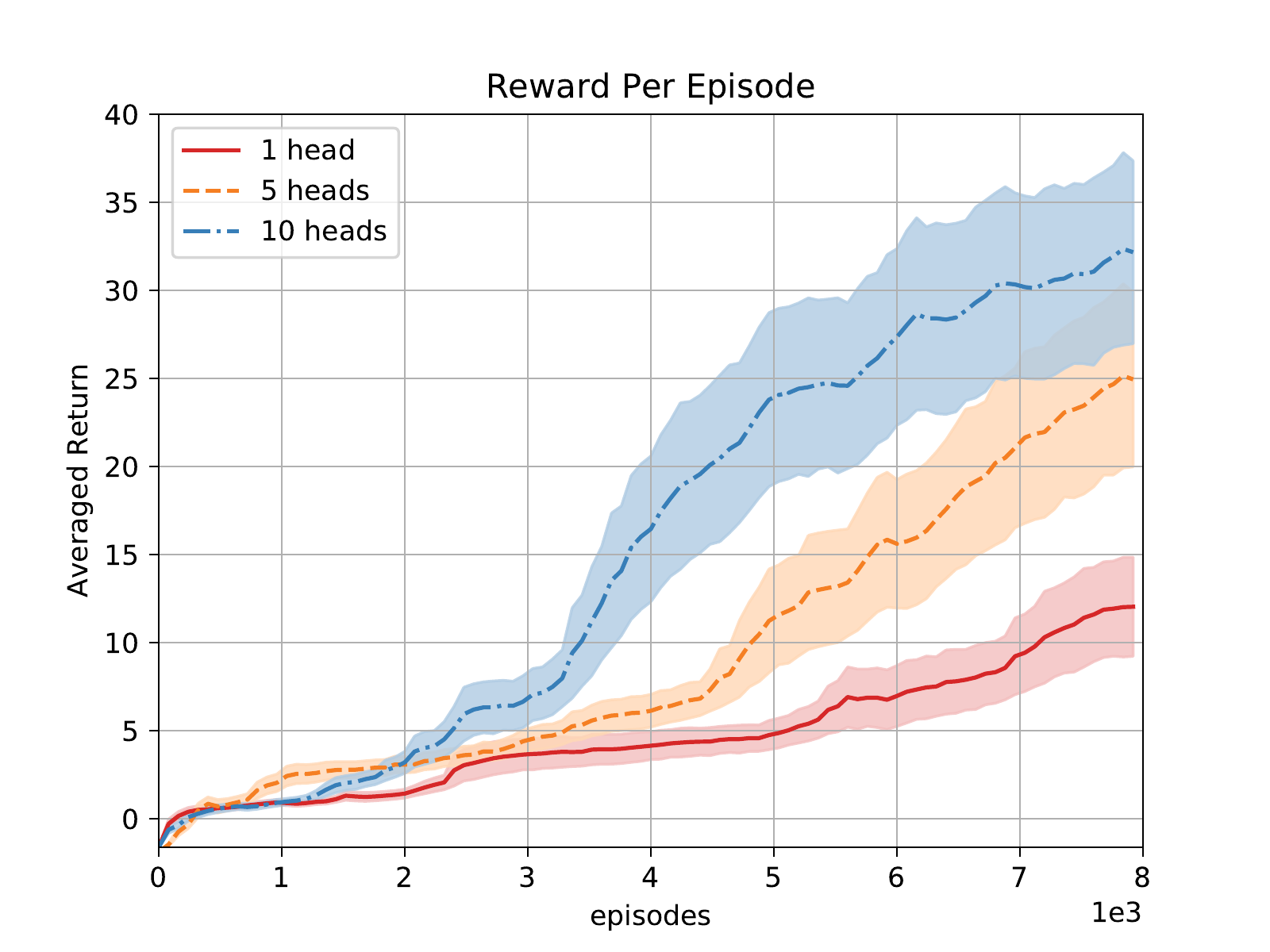}
	\caption{Episode reward comparison among three modifications of  DB-DDPG with different number of bootstrapped heads. Single-head model could be regarded as vanilla DDPG. Multi-head models scored higher rewards than single-head model.}\label{fig:fig-result}
\end{figure}

We evaluated the algorithm on a single machine. Results in Figure~\ref{fig:fig-result} showed that multi-head architecture brought significant performance increase. These models with different multiple heads were pre-trained using previous experience memory in the warm-up phase before training. The single-head model could be regarded as vanilla DDPG. These agents were trained using the same techniques and hyperparameters mentioned above following Algorithm~\ref{code-dbddpg}. The figure showed that DB-DDPGs outperformed vanilla DDPG on performance ensuring faster training and stability. The model with 10-head scored the highest cumulative reward by more efficient exploration and ensemble training.

\subsection{Discussion}
We proposed the \textit{Double Bootstrapped DDPG} algorithm for high-dimensional continuous control tasks. It was demonstrated that DB-DDPG with multiple heads led to significantly faster learning than vanilla DDPG while retaining stability in the Learning to Run challenge.
	
\noindent
Going forward, DB-DDPG can be not only efficient on a single machine, but parallelizable up to more machines boosting the learning. Furthermore, we believe this architecture is also available for dealing with multi-task reinforcement learning problems, which means each head only concentrates on its own sub-task when training then makes its own professional decision for ensemble collaboratively.

\begin{algorithm}[htbp]
	\caption{Double Bootstrapped DDPG}\label{code-dbddpg}
	\begin{algorithmic}
		\State \textbf{Input:} head number $K$, mini-batch size $N$, maximum training episode $E$
		\State \textbf{Initialize:} Randomly initialize critic network $Q(s,a|\theta^Q)$ with $K$ outputs $\{{Q_k}\}_{k=1}^K$, and actor network $\mu(s|\theta^\mu)$ with $K$ outputs $\{{\mu_k}\}_{k=1}^K$.
		\State Initialize critic target networks $\theta_{1}^{Q'},...,\theta_{K}^{Q'}$ with weights $\theta_{k}^{Q'}\leftarrow\theta_{k}^{Q}$ and actor target networks $\theta_{1}^{\mu'},...,\theta_{K}^{\mu'}$ with weights $\theta_{k}^{\mu'}\leftarrow\theta_{k}^{\mu}$
		\State Initialize replay buffer $R$, masking distribution $M$
		\For{episode $e = 1, E$}
		\State Initialize a random process $\mathcal{N}$ for action exploration
		\State Receive initial observation state $s_0$
		\State Pick a pair of activated critic and actor networks using $k{\sim}Uniform\{1,...,K\}$
		\For{step $t = 1, T$}
		\State Select action $a_t$ from actions $\{a|a^k=\mu(s_t|\theta_{k}^{\mu})\}_{k=1}^K$ according to the policies and exploration noise as following,
		\State{} $\vcenter{
			\begin{equation*}
			a_t=\argmax_a\{\sum_{i=1}^{K}Q_{i}(s_t,a)|_{{a=\mu_k(s_t)}}\}_{k=1}^K+\mathcal{N}_t
			\end{equation*}
		}$
		\State Execute action $a_t$ then observe reward $r_t$ and new state $s_{t+1}$
		\State Sample bootstrapped mask $m_t{\sim}M$
		\State Store transition $(s_t,a_t,r_t,s_{t+1},\textbf{\textit{m}}_t)$ in $R$
		\State Sample a random minibatch of $m$ transitions $(s_i,a_i,r_i,s_{i+1},m_i)$
		\State Set $y_i=r_i+{\gamma}Q'(s_{i+1},\mu'(s_{i+1}|\theta_{k}^{\mu'})|\theta_{k}^{Q'})$
		\State Update critic network by minimizing the loss: $L=\frac1N\sum_im_{i}^{k}(y_i-Q(s_i,a_i|\theta_{k}^Q))^2$
		\State Update actor network using the sampled policy gradient: 
		\State{} $\vcenter{
			\begin{equation*}
			\nabla_{\theta_{k}^\mu}\approx\frac1N{\sum}_i\nabla_aQ(s,a|\theta_{k}^Q)|_{s=s_i, a=\mu(s_i)}\nabla_{\theta_{k}^\mu}\mu(s|\theta_{k}^\mu)|_{s=s_i}
			\end{equation*}
		}$
		\State Update the target networks:
		\State{} $\vcenter{
			\begin{equation*}
			\theta_{k}^{Q'}\leftarrow\tau\theta_{k}^{Q}+(1-\tau)\theta_{k}^{Q'}
			\end{equation*}
		}$
		\State{} $\vcenter{
			\begin{equation*}
			\theta_{k}^{\mu'}\leftarrow\tau\theta_{k}^{\mu}+(1-\tau)\theta_{k}^{\mu'}
			\end{equation*}
		}$
		\EndFor
		\EndFor
	\end{algorithmic}
\end{algorithm}

\section{Plain DDPG with soft target networks}\label{s:anton}
\sectionauthor{Anton Pechenko}

We used DDPG with soft target networks and Ornstein-Uhlenbeck process. Discount factor was 0.99, replay buffer was 1 000 000, batch size was 64. For training we used 7 simulators which was run in parallel.

\subsection{Method}

\subsubsection{State description}
To remove redundancy from the state space and make training more efficient it is often reasonable to exploit state space symmetry. That is why we used \textit{first person view} transformation of the observation vector. That means we subtracted coordinates and angle of pelvis from others bones coordinates and angles. Also we assigned zero to $X$ coordinate of the pelvis to collapse observation space along $X$ coordinate to make run performance independent of $X$ coordinate. We had no information about ground reaction forces in observation vector. Since it seemed very important for running it was needed to estimate it. To that end, we constructed a state vector from a sequence of the three last observation vectors.

\subsubsection{Training process}
We used two separated multilayer perceptrons \citep{Rumelhart:1986:LIR:104279.104293} for actor and critic with 5 hidden layers, 512 neurons each. In three out of seven agents we random 20-40 actions at start and then started collecting $(S, A, R, S')$ experience items, i.e. a previous state state, an action, a reward and a result state. This increases variety of starting points and increases quality of replay buffer experience. For 30\% of simulations we turned off obstacles in order to increase the variety of replay buffer. Training consisted of two steps. The first step was optimization with Adam \citep{DBLP:journals/corr/KingmaB14} with $1e-4$ learning rate till agents increase its score at maximum, which, in our case, turned out to be approximately 37 meters. In the second step, we trained the network with the stochastic gradient descent (SGD) using the learning rate of  $5e-5$. During the SGD step agents decreased running velocity but increase robustness to falls and obstacles overcoming resulting in 26 meters score.

\subsection{Experiments and results}

The implementation of this solution can be found as a part of the RL-Server software available at github\footnote{ \url{https://github.com/parilo/rl-server}}. The RL-Server is a python application using Tensorflow \citep{tensorflow2015-whitepaper}, with DQN \citep{DBLP:journals/corr/MnihKSGAWR13} and DDPG algorithms included. The entire training process, including replay buffer, training and inference steps is performed within the RL-Server application. A lightweight client code can be included in parallel running environments. It enables multiple languages thanks to an open communication protocol.

\section{PPO with reward shaping}\label{s:adam}
\sectionauthor{Adam Stelmaszczyk and Piotr Jarosik}

We trained our final model with PPO on 80 cores in 5 days using reward shaping, extended and normalized observation vector. We recompiled OpenSim with lower accuracy to have about 3x faster simulations.

\subsection{Methods}

Our general approach consisted of two phases: exploration of popular algorithms and exploitation of the most promising one. In the first phase we tried 3 algorithms (in the order we tried them): 
Deep Deterministic Policy Gradient (DDPG) (keras-rl implementation \cite{plappert}),
Proximal Policy Optimization (PPO) \cite{schulman} (OpenAI baselines implementation \cite{baselines}), Evolution Strategies (ES) \cite{salimans} (OpenAI implementation \cite{starter}).
We also tried 3 improvements:
changing the reward function during training (reward shaping), improving observations (feature engineering) and normalizing observations.

We started our experiments with DDPG (without improvements) and we could not achieve good results. That was probably because of the bad normalization or not enough episodes. We had problems parallelizing keras-rl and we were using only one process. Therefore, we switched to PPO and ES, for which learning plots and parallelization looked better. We were incrementally adding improvements to these two techniques. In the end we used the default hyperparameters for all the algorithms, their values can be found in the full source code \cite{stelmaszczyk}.

\subsubsection{Reward shaping}

We guided learning to promising areas by shaping the reward function. We employed: a penalty for straight legs, a penalty for \verb|pelvis.x| greater than \verb|head.x| (causing a skeleton to lean forward), adding $0.01$ reward for every time step (to help take the first step and get out of local maximum) and using velocity instead of distance passed, found in \cite{heess2017emergence}. Using velocity rewards passing the same distance in less time steps.
   
\subsubsection{Feature engineering}

We changed all \verb|x| positions from absolute to relative to \verb|pelvis.x|. Thanks to that similar poses were represented by similar observations. We also extended the observation vector from $41$ values to $82$ by adding: the remaining velocities and accelerations of the body parts, ground touch indicator for toes and talus, a queue of two obstacles: the next one (preserved from the original observation) and the previous one. Without this, when passing over an obstacle, agent would lose sight of the obstacle underneath it as it would immediately switch to the next one.

\subsubsection{Normalizing observations}

We logged the mean and standard deviation of all the observations to see if the normalization was done correctly. By default, baselines PPO implementation used a filter which automatically normalized every observation. For every observation, it was keeping its running mean and standard deviation. It did normalization by subtracting the mean and dividing by std. This worked well for most of the observations, however for some it was problematic.
For constants, e.g. the strength of psoas, the standard deviation would be 0. The code in that case was just passing this observation as it is. The magnitude of an observation was treated as an importance when passed to a network. Too big values would saturate all the other smaller inputs (which may be more important).
Also, the first strength of psoas had a different value than the following ones (due to a bug in the challenge environment, later fixed). So, the filter would calculate some arbitrary mean with standard deviation and later use them.
Another problem was that some observations were most of the time were close to 0, but were shooting up in some moments to greater values, e.g. velocity. This resulted in huge values (because initial standard deviation was close to 0), saturating the network.

Because all of these problems, we skipped the auto normalizing and manually normalized every observation. Iteratively, we were running our model and visualizing mean with standard deviation for all the observations. Then we were correcting the normalization of observations which mean was far from 0 or standard deviation far from 1.

\begin{figure}
	 
    \begin{subfigure}[t]{1\textwidth}
	\centering
      \includegraphics[scale=.44]{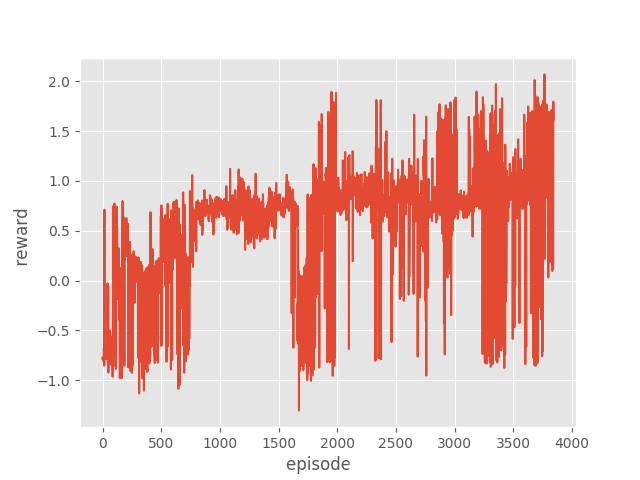}
      \caption{DDPG}
    \end{subfigure}
    \centering
	\begin{subfigure}[t]{.45\textwidth}
      \centering
      \includegraphics[scale=.52]{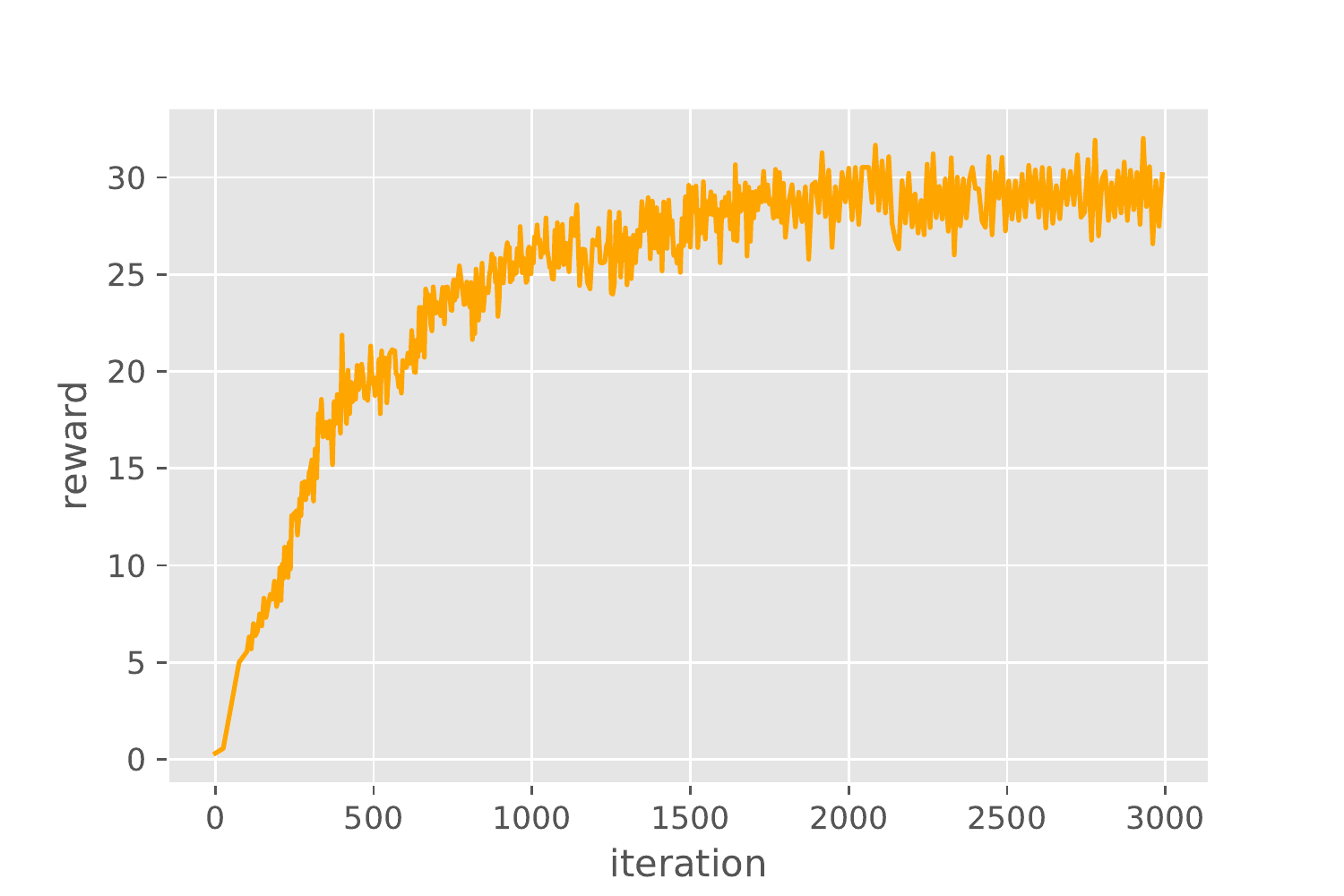}
       \caption{PPO}
    \end{subfigure}
    \begin{subfigure}[t]{.45\textwidth}
    \centering
      \includegraphics[scale=.5]{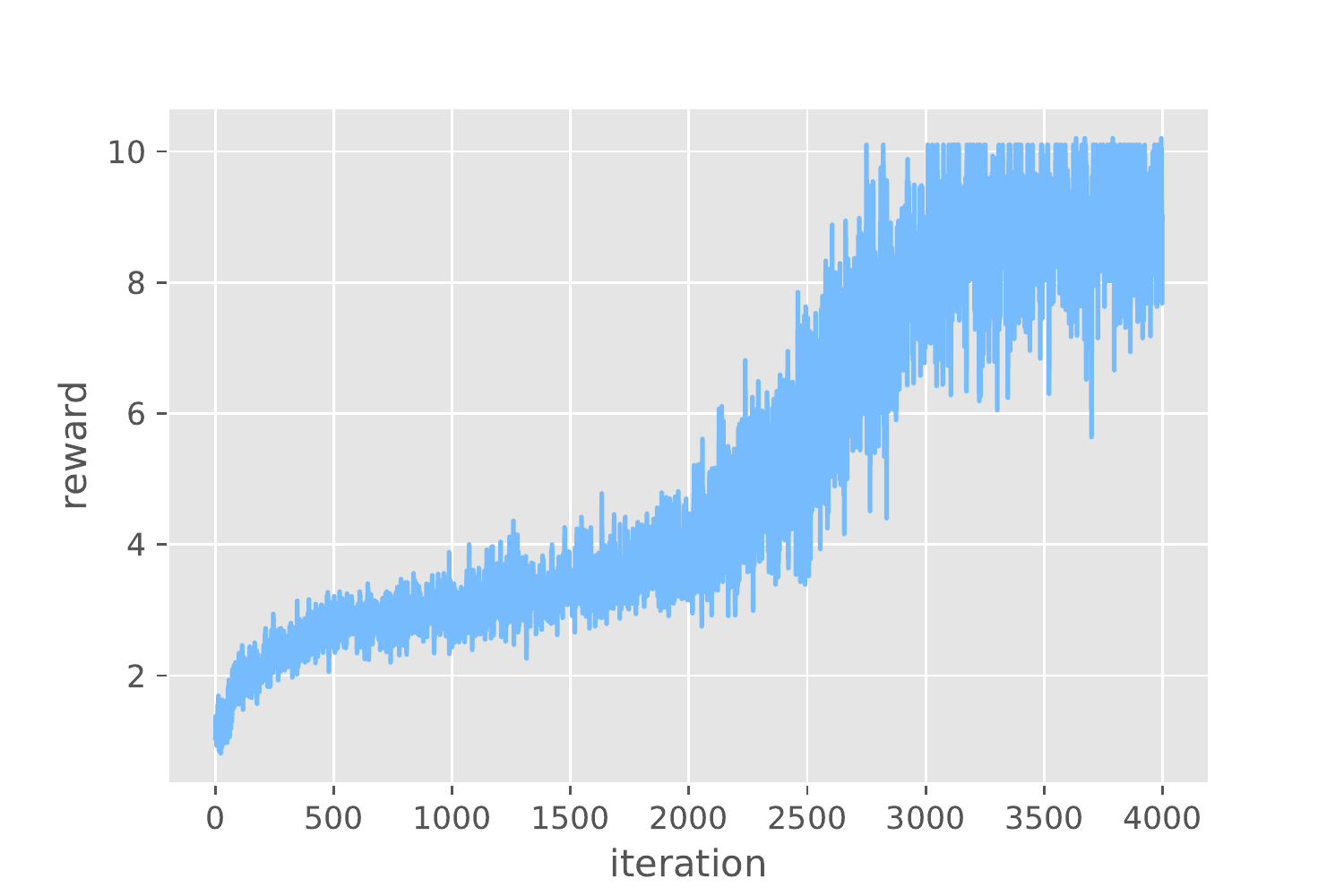}
      \label{fig:es}
      \caption{ES}
    \end{subfigure}
    
    \caption{Mean rewards (after reward shaping, so higher than during grading) achieved by DDPG, PPO and ES in one training run. The x axis in the DDPG plot stands for episodes, in PPO and ES - for iterations (multiple episodes), because of the different implementations used (keras-rl for DDPG and OpenAI baselines for PPO \& ES). The mean reward of DDPG was very variable. We experienced the most stable learning with PPO. ES gave us worse results, after 3000 iterations the skeleton stayed still for 1000 time steps, scoring around 10, because we rewarded 0.01 for every survived time step.}
    \label{meanreward}
\end{figure}

\subsection{Experiments and results}

We found that OpenSim simulator was the bottleneck, accounting for about 99\% of the CPU  time. To speed it up about 3x, we changed the simulator's accuracy from default 0.1\% to 3\%, following \url{https://github.com/ctmakro/stanford-osrl#the-simulation-is-too-slow}. Any change made in our environment could have introduced a bias when grading on the server, however we didn't notice any significant score changes.

We conducted our experiments on:
\emph{Chuck}, an instance with 80 CPU cores (Intel Xeon), provided by Faculty of Mathematics, Informatics, and Mechanics, University of Warsaw.
\emph{Devbox}, an instance with 16 CPU and 4 CUDA GPU cores, provided by Institute of Fundamental Technological Research, Polish Academy of Sciences.
AWS c5.9xlarge and c4.8xlarge instances, in the last 4 days, sponsored by Amazon.

We checked OpenSim simulator performance on Nvidia GPU cores (with NVBLAS support), however it didn't reduce the computation time. Final training was done on \emph{Chuck} due to its large number of CPU cores. Different training runs were often resulting in very different models, some of which were stuck in local maxima early. Because of that, we started 5 to 10 separate training runs. We monitored their plots and visualized the current policy from time to time. If we judged that the model was not promising - e.g. it was stuck in local maxima or looked inferior to some other one - we stopped it and gave the resources to the more promising trainings.

In Figure \ref{meanreward} we present mean episode reward obtained by DDPG, PPO and ES during one training run. We achieved the best result using PPO, our score in the final stage was 18.63 meters. The final score was taken as the best score out of 5 submissions. The remaining scores were: 18.53, 16.14, 15.19, 14.5. The average of our 5 final submissions was 16.6.

\subsection{Discussion}

The average score during DDPG training was fluctuating a lot, sometimes it could also drop and never regain. We tried to tune the hyperparameters, without any gain though. Our problems were most probably due to bad data normalization or not enough episodes. The average score with PPO was not fluctuating as much and did not suddenly drop. That is why we switched to PPO and stayed with it until the end of the competition. ES usage in the Learning to Run environment should be more thoroughly examined.

There are a few things we would do differently now. We would try DDPG OpenAI baselines implementation. We would use simpler and well-known environment in the beginning, e.g. Walker2d and reproduce the results. We would make sure normalization is done correctly. We would try Layer Normalization instead of tedious manual normalization. We would tune hyperparameters in the end and in a rigorous way \cite{henderson}. We would repeat an action $n$ times (so-called \textit{frame skip}). We would learn also on mirror flips of observations as shown in \cite{pavlov}. Finally, we would use TensorBoard or similar for visualizations.

\section{Leveraging emergent muscles-activation patterns: from DDPG in a continuous action space to DQN with a set of actions}\label{s:citec}
\sectionauthor{Andrew Melnik, Malte Schilling and Helge Ritter}

A continuous action space with a large number of dimensions allows for complex manipulations. However, often, only a limited number of points in the action space is used. Furthermore, approaches like Deep Deterministic Policy Gradient (DDPG) may stick to a local optimum, where different optima have different sets of points in use. Therefore, to generalize over several local optima, we collected such points in the action space from different local optima and leveraged them in Deep Q-Network (DQN) \cite{mnih2015human} learning.

\subsection{Methods}

\begin{figure}[h!]
  \centering
  \includegraphics[width=0.96\textwidth]{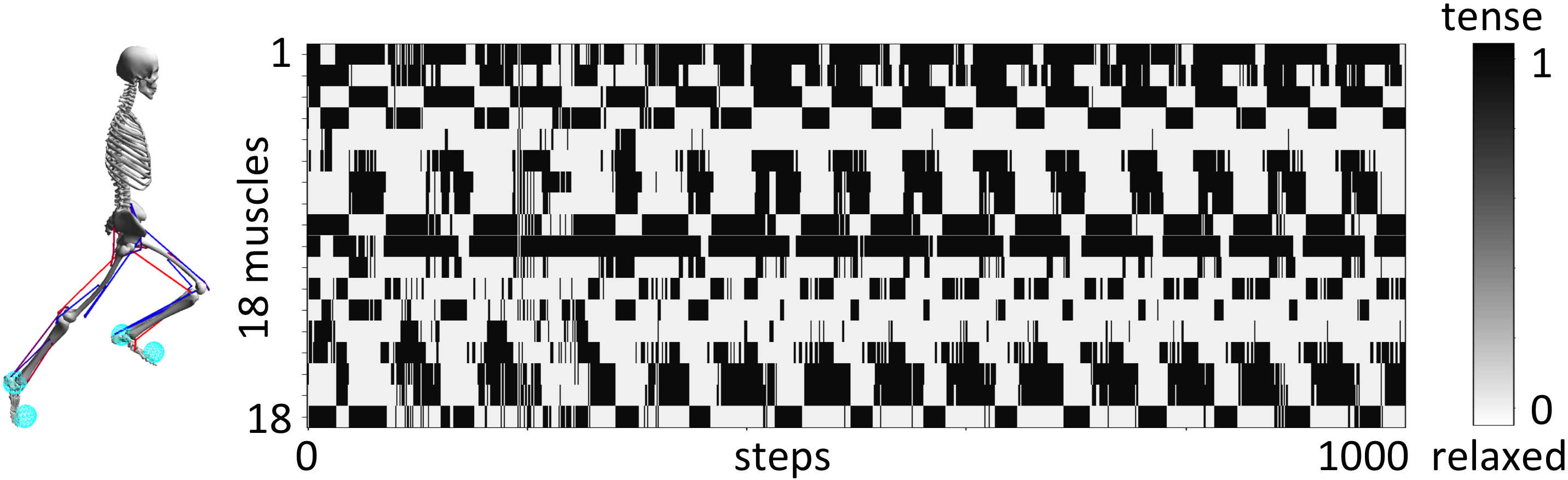}
  \caption{Muscle activations in an episode. Columns are actor NN outputs after training.} 
  \label{fig:bwPatterns}
\end{figure}

Our approach consisted of two parts. In the first part, we applied the DDPG model to the Learning to Run environment. Our model consisted of actor and critic sub-networks with parameters used as recommended by the Getting Started guide\footnote{\url{https://github.com/stanfordnmbl/osim-rl}}. For the initial exploration of the continuous 18-dimensional action space, we used Ornstein-Uhlenbeck (OU) process \cite{uhlenbeck1930theory} to generate temporally correlated noise which we added to the actor Neural Network (NN) output values. The model learned reliable muscles-activation patterns to perform successful locomotion (Fig.~\ref{fig:bwPatterns}). After training, when the agent reached a performance plateau, the outputs of the actor NN became either equal to 0 or 1 (vectors of 18 binary values). We let the agents run further and collected a set of actor NN outputs (Table.~\ref{table:patterns}, Fig.~\ref{fig:Pareto}). To generalize over many successful locomotion strategies, we collected patterns from different independently trained agents.

\begin{wrapfigure}[12]{R}{0.6\textwidth}
\centering
\includegraphics[width=0.6\textwidth]{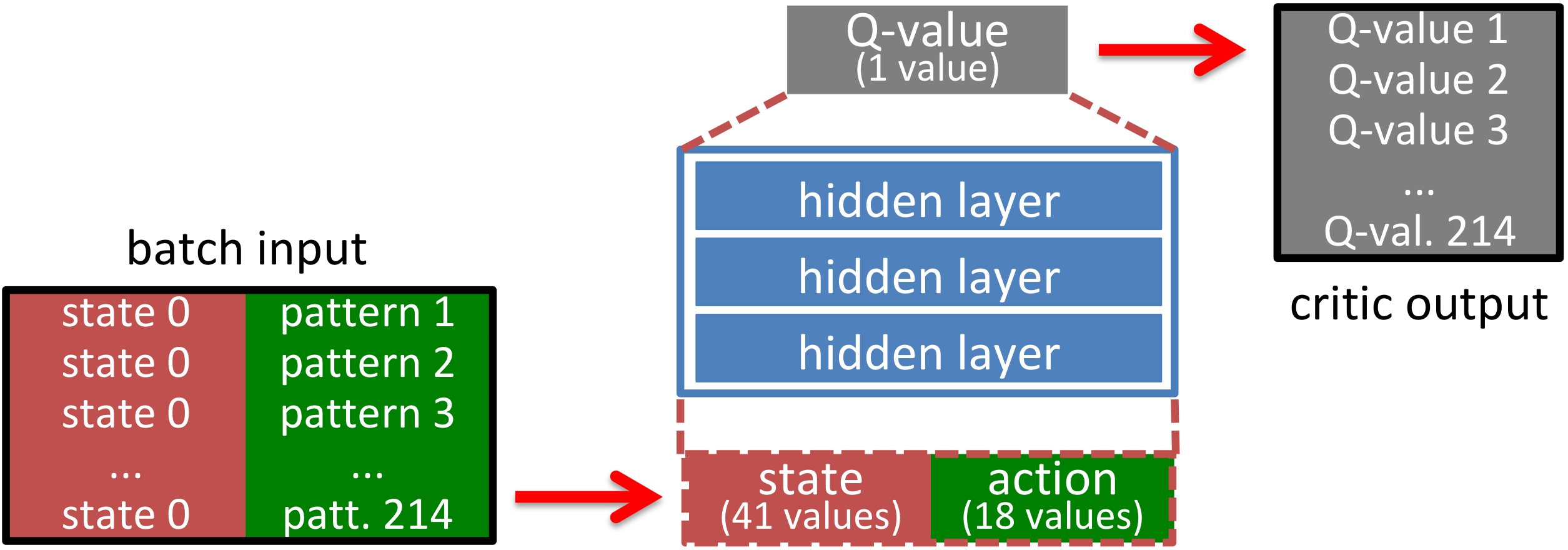}
\caption{\label{fig:while}Q-value estimation for a set of collected patterns.}
\end{wrapfigure}

\begin{figure}[h!]
  \centering
  \includegraphics[width=0.9\textwidth]{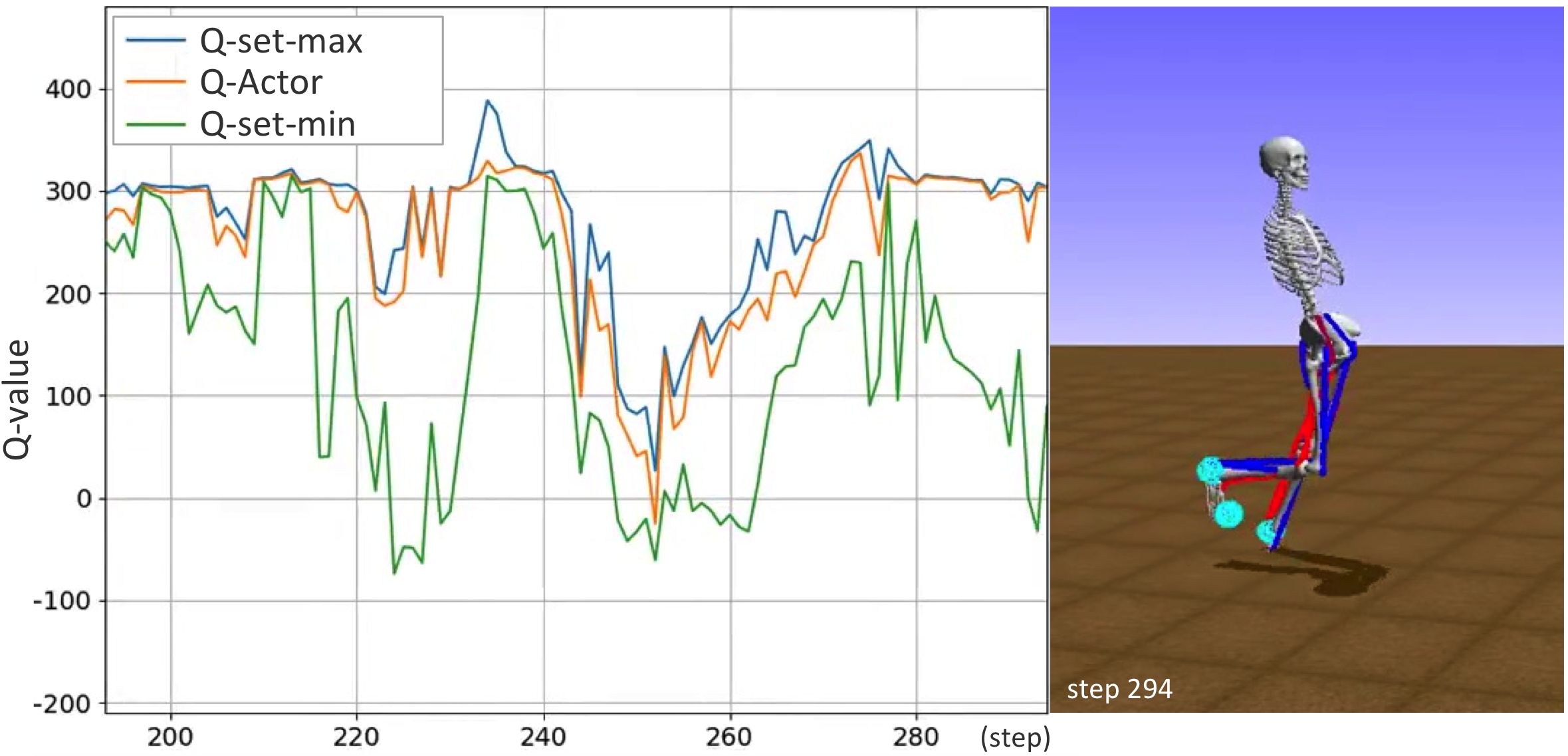}
  \caption{Q-value range. Upper and lower curves represent maximum and minimum Q-values for the set of collected muscles-activation patterns in given states. The middle curve represents Q-values of the muscles-activation patterns proposed by the actor NN.}
  \label{fig:qgraph}
\end{figure}

In the second part, we used the set of collected patterns in two different scenarios. In the first scenario, we continued with training of DDPG models and used Q-values of the set of collected patterns for exploration, as, according to the critic NN, actor's outputs were close, but not equal to the highest Q-values of the set of collected patterns (Fig. ~\ref{fig:qgraph}). The collected set allowed us to vastly reduce the action-search-space to a moderate set of useful muscles-activation patterns. Exploration by selection of patterns with highest Q-values allowed, in many cases, to increase the score further by 10-20\%, after the DDPG agent reached a plateau of performance. In the second scenario, following the DQN approach, we used solely the critic NN (Fig.~\ref{fig:while}) to train new agents. To get Q-values for the set of collected muscles-activation patterns we concatenated them with state values and fed the batch to the critic NN (Fig.~\ref{fig:while}).

\begin{table}[h!]
  \begin{varwidth}[b]{0.49\linewidth}
    \centering
    \scalebox{0.82}{
    \begin{tabular}{p{1.cm}p{3.5cm}p{1.cm}p{1.cm}}
      \hline
      Patt. ID & Muscles-Activation \newline Pattern & Occur. freq.(\%) & Sum (\%) \\
      \hline
      1	& 110100001101001110	& 8.1	& 8.1 \\
      2	& 110100001100000110	& 4.5	& 12.6 \\
      3	& 000100001101001110	& 4.0	& 16.6 \\
      4	& 111001110110100001	& 3.5	& 20.1 \\
      5	& 101001110110100001	& 3.5	& 23.6 \\
      6	& 111001101100000110	& 2.7	& 26.3 \\
      7	& 110100001100011110	& 2.4	& 28.7 \\
      8	& 010100001101001110	& 2.3	& 31.0 \\
      ...	& ...	& ...	& ... \\
      214	& 000000001101001001	& 0.1	& 100.0 \\
      \hline
    \end{tabular}
    }
    \vspace*{2mm}
    \caption{Patterns of muscles activation and occurrence frequencies.}
    \label{table:patterns}
  \end{varwidth}%
  \hfill
  \begin{minipage}[b]{0.45\linewidth}
    \centering
    \includegraphics[width=60mm]{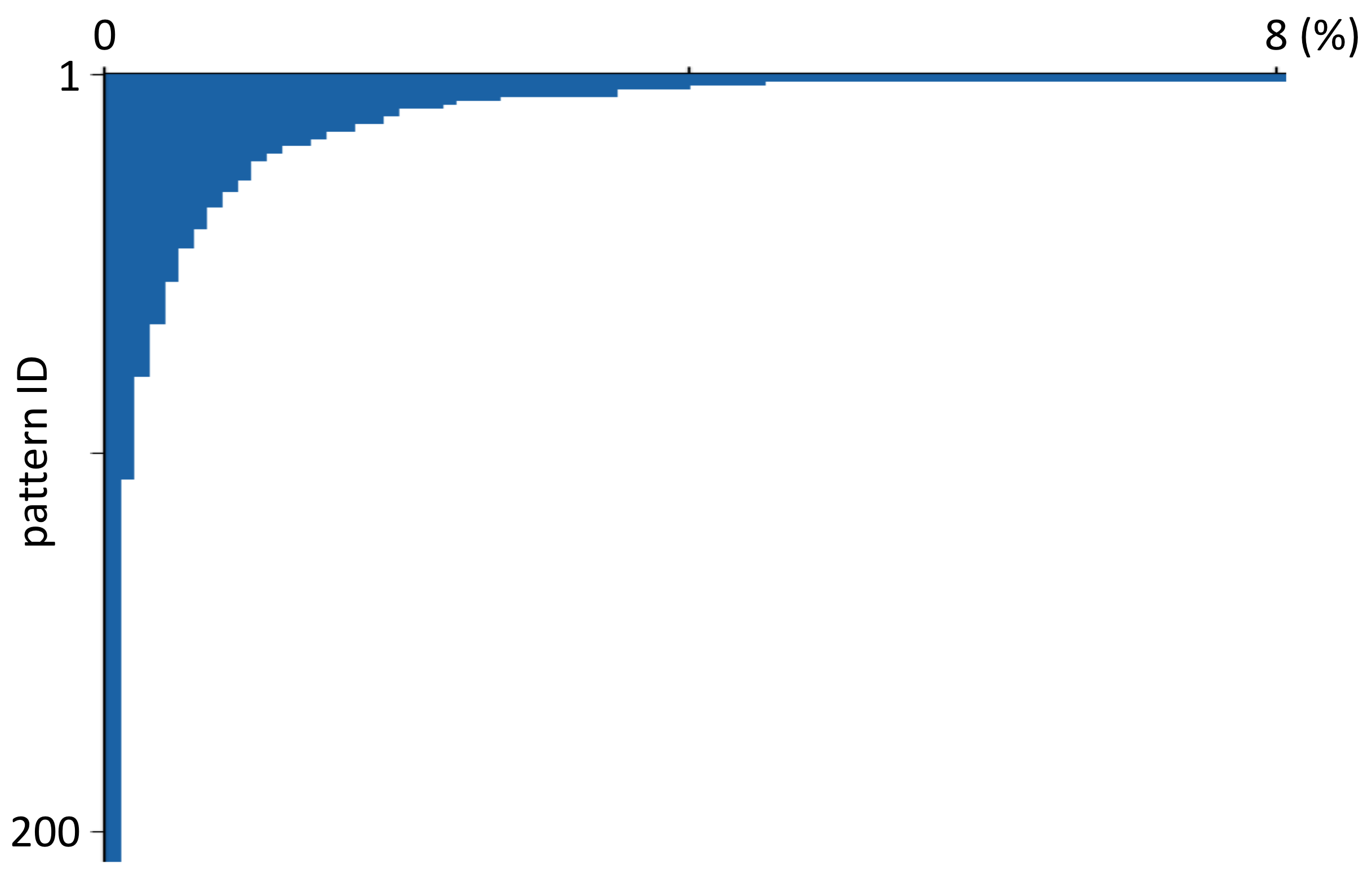}
    \vspace*{2mm}
    \captionof{figure}{Pareto distribution of occurrence frequencies of the muscles-activation patterns.}
    \label{fig:Pareto}
  \end{minipage}
\end{table}

While a binary vector of 18 values would have $2^{18}=262144$ potential combinations, we found only a set of 214 muscles-activation patterns for trained models. That reduced exploration space to a moderate set of meaningful patterns. Certain patterns occurred more frequently than other (Table~\ref{table:patterns}) with the 8 most frequent patterns representing already more than 30 \% of all executed actions. About a half of the collected patterns (108 patterns) occurred only once (per episode of 1000 steps). Occurrence frequencies of the collected patterns (Table~\ref{table:patterns}) demonstrated a Pareto distribution (Fig.~\ref{fig:Pareto}).

\subsection{Discussion}
We presented alternative (DQN based) approaches to explore and select actions in the continues action space. However, the prerequisite is to have a set of meaningful actions for the task. Exploiting the set of collected muscles-activation patterns for further training has shown to lead to better performance and overall we want to consider in the future how to bootstrap the information further from emerging activation patterns and updating this collection. As a possible extension, we could take into account the reflection symmetry of muscles-activation patterns for the left and right legs. The set of 214 unique muscles-activation patterns for 18 muscles (two legs) contains only a set of 56 unique muscles-activation patterns for 9 muscles (per leg). In a way, this work is related to the ideas of hierarchical reinforcement learning \cite{heess2016} and the work on learning Dynamic Movement Primitives \cite{Schaal2006,Ijspeert_NC_2013} which are attractor systems of a lower dimensionality on the lower levels of such hierarchical systems.

\section{Affiliations and acknowledgments}\label{s:acknowledgements}

\textbf{Organizers:} {\L}ukasz Kidzi\'nski, Carmichael Ong, Jennifer Hicks and Scott Delp are affiliated with Department of Bioengineering, Stanford University. Sharada Prasanna Mohanty, Sean Francis and Marcel Salathé are affiliated with Ecole Polytechnique Federale de Lausanne. Sergey Levine is affiliated with University of California, Berkeley.

\textbf{Team PKU (place 2nd, Section \ref{s:pku}):} Zhewei Huang and Shuchang Zhou are affiliated with Bejing University.
\textbf{Team reason8.ai (place 3rd, Section \ref{s:reason8}):} Mikhail Pavlov, Sergey Kolesnikov and Sergey Plis are affiliated with reason8.ai.
\textbf{Team IMCL (place 4th, Section \ref{s:imcl}):} Zhibo Chen, Zhizheng Zhang, Jiale Chen and Jun Shi are affiliated with Immersive Media Computing Lab, University of Science and Technology of China.
\textbf{Team deepsense.ai (place 6th, Section \ref{s:deepsense}):} Henryk Michalewski is affiliated with Institute of Mathematics, Polish Academy of Sciences and deepsense.ai. Piotr Miłoś and Błażej Osiński are affiliated with  Faculty of Mathematics, Informatics, and Mechanics, University of Warsaw and deepsense.ai.
\textbf{Team THU-JGeek (place 8th, Section \ref{s:jgeek}):} Zhuobin Zheng, Chun Yuan and Zhihui Lin are affiliated with Tunghai University.
\textbf{Team Anton Pechenko (place 16th, Section \ref{s:anton}):} Anton Pechenko is affiliated with Yandex.
\textbf{Team Adam Stelmaszczyk (place 22nd, Section \ref{s:adam}):} Adam Stelmaszczyk is affiliated with Faculty of Mathematics, Informatics, and Mechanics, University of Warsaw. Piotr Jarosik is affiliated with Institute of Fundamental Technological Research, Polish Academy of Sciences. 
\textbf{Team Andrew Melnik (place 28th, Section \ref{s:citec}):} Andrew Melnik, Malte Schilling and Helge Ritter are affiliated with CITEC, Bielefeld University.

Team deepsense.ai was supported by the PL-Grid Infrastructure:
Prometheus and Eagle supercomputers, located respectively in the
Academic Computer Center Cyfronet at the AGH University of Science and
Technology in Kraków and the Supercomputing and Networking Center in
Poznan. The deepsense.ai team also expresses gratitude to NVIDIA and
Goodeep for providing additional computational resources used during
the experiment.

The challenge was co-organized by the Mobilize Center, a National Institutes of Health Big Data to Knowledge (BD2K) Center of Excellence supported through Grant U54EB020405. The challenge was partially sponsored by Nvidia who provided DGX Station™ for the first prize in the challenge, and GPUs Titan V for the second and the third prize, by Amazon Web Services who provided 30000 USD in cloud credits for participants, and by Toyota Research Institute who funded one travel grant.


\begin{thebibliography}{10}
\providecommand{\url}[1]{{#1}}
\providecommand{\urlprefix}{URL }
\expandafter\ifx\csname urlstyle\endcsname\relax
  \providecommand{\doi}[1]{DOI~\discretionary{}{}{}#1}\else
  \providecommand{\doi}{DOI~\discretionary{}{}{}\begingroup
  \urlstyle{rm}\Url}\fi

\bibitem{tensorflow2015-whitepaper}
Abadi, M., Agarwal, A., Barham, P., Brevdo, E., Chen, Z., Citro, C., Corrado,
  G.S., Davis, A., Dean, J., Devin, M., Ghemawat, S., Goodfellow, I., Harp, A.,
  Irving, G., Isard, M., Jia, Y., Jozefowicz, R., Kaiser, L., Kudlur, M.,
  Levenberg, J., Man\'{e}, D., Monga, R., Moore, S., Murray, D., Olah, C.,
  Schuster, M., Shlens, J., Steiner, B., Sutskever, I., Talwar, K., Tucker, P.,
  Vanhoucke, V., Vasudevan, V., Vi\'{e}gas, F., Vinyals, O., Warden, P.,
  Wattenberg, M., Wicke, M., Yu, Y., Zheng, X.: {TensorFlow}: Large-scale
  machine learning on heterogeneous systems (2015).
\newblock \urlprefix\url{https://www.tensorflow.org/}.
\newblock Software available from tensorflow.org

\bibitem{anonymous2018distributional}
Anonymous: Distributional policy gradients.
\newblock International Conference on Learning Representations  (2018).
\newblock \urlprefix\url{https://openreview.net/forum?id=SyZipzbCb}

\bibitem{ba2016layer}
Ba, J.L., Kiros, J.R., Hinton, G.E.: Layer normalization.
\newblock arXiv preprint arXiv:1607.06450  (2016)

\bibitem{cloning}
Bratko, I., Urbančič, T., Sammut, C.: Behavioural cloning: Phenomena, results
  and problems.
\newblock IFAC Proceedings Volumes \textbf{28}(21), 143 -- 149 (1995).
\newblock \doi{https://doi.org/10.1016/S1474-6670(17)46716-4}.
\newblock
  \urlprefix\url{http://www.sciencedirect.com/science/article/pii/S1474667017467164}

\bibitem{clevert2015fast}
Clevert, D.A., Unterthiner, T., Hochreiter, S.: Fast and accurate deep network
  learning by exponential linear units (elus).
\newblock arXiv preprint arXiv:1511.07289  (2015)

\bibitem{baselines}
Dhariwal, P., Hesse, C., Plappert, M., Radford, A., Schulman, J., Sidor, S.,
  Wu, Y.: {OpenAI Baselines}.
\newblock \url{https://github.com/openai/baselines} (2017)

\bibitem{dietterich2000ensemble}
Dietterich, T.G., et~al.: Ensemble methods in machine learning.
\newblock Multiple classifier systems \textbf{1857}, 1--15 (2000)

\bibitem{robot_shaping}
Dorigo, M., Colombetti, M.: Robot Shaping: An Experiment in Behavior
  Engineering.
\newblock MIT Press, Cambridge, MA, USA (1997)

\bibitem{heess2017emergence}
Heess, N., Sriram, S., Lemmon, J., Merel, J., Wayne, G., Tassa, Y., Erez, T.,
  Wang, Z., Eslami, A., Riedmiller, M., et~al.: Emergence of locomotion
  behaviours in rich environments.
\newblock arXiv preprint arXiv:1707.02286  (2017)

\bibitem{heess2016}
Heess, N., Wayne, G., Tassa, Y., Lillicrap, T.P., Riedmiller, M.A., Silver, D.:
  Learning and transfer of modulated locomotor controllers.
\newblock CoRR \textbf{abs/1610.05182} (2016).
\newblock \urlprefix\url{http://arxiv.org/abs/1610.05182}

\bibitem{henderson}
{Henderson}, P., {Islam}, R., {Bachman}, P., {Pineau}, J., {Precup}, D.,
  {Meger}, D.: {Deep Reinforcement Learning that Matters}.
\newblock ArXiv e-prints  (2017)

\bibitem{hessel2017rainbow}
Hessel, M., Modayil, J., Van~Hasselt, H., Schaul, T., Ostrovski, G., Dabney,
  W., Horgan, D., Piot, B., Azar, M., Silver, D.: Rainbow: Combining
  improvements in deep reinforcement learning.
\newblock arXiv preprint arXiv:1710.02298  (2017)

\bibitem{Ijspeert_NC_2013}
Ijspeert, A., Nakanishi, J., Pastor, P., Hoffmann, H., Schaal, S.: Dynamical
  movement primitives: Learning attractor models for motor behaviors.
\newblock Neural Computation \textbf{25}, 328--373 (2013).
\newblock
  \urlprefix\url{http://www-clmc.usc.edu/publications/I/ijspeert-NC2013.pdf}.
\newblock Clmc

\bibitem{jaskowski2018rltorunfast}
Ja\'skowski, W., Lykkeb{\o}, O.R., Toklu, N.E., Trifterer, F., Buk, Z.,
  Koutn\'{i}k, J., Gomez, F.: {Reinforcement Learning to Run... Fast}.
\newblock In: S.~Escalera, M.~Weimer (eds.) NIPS 2017 Competition Book.
  Springer, Springer (2018)

\bibitem{ji20133d}
Ji, S., Xu, W., Yang, M., Yu, K.: 3d convolutional neural networks for human
  action recognition.
\newblock IEEE transactions on pattern analysis and machine intelligence
  \textbf{35}(1), 221--231 (2013)

\bibitem{kidzinski2018learningtorun}
Kidzi\'nski, {\L}., Sharada, M.P., Ong, C., Hicks, J., Francis, S., Levine, S.,
  Salath\'e, M., Delp, S.: Learning to run challenge: Synthesizing
  physiologically accurate motion using deep reinforcement learning.
\newblock In: S.~Escalera, M.~Weimer (eds.) NIPS 2017 Competition Book.
  Springer, Springer (2018)

\bibitem{DBLP:journals/corr/KingmaB14}
Kingma, D.P., Ba, J.: Adam: {A} method for stochastic optimization.
\newblock CoRR \textbf{abs/1412.6980} (2014).
\newblock \urlprefix\url{http://arxiv.org/abs/1412.6980}

\bibitem{klambauer2017self}
Klambauer, G., Unterthiner, T., Mayr, A., Hochreiter, S.: Self-normalizing
  neural networks.
\newblock arXiv preprint arXiv:1706.02515  (2017)

\bibitem{lillicrap2015continuous}
Lillicrap, T.P., Hunt, J.J., Pritzel, A., Heess, N., Erez, T., Tassa, Y.,
  Silver, D., Wierstra, D.: Continuous control with deep reinforcement
  learning.
\newblock arXiv preprint arXiv:1509.02971  (2015)

\bibitem{DBLP:journals/corr/MnihKSGAWR13}
Mnih, V., Kavukcuoglu, K., Silver, D., Graves, A., Antonoglou, I., Wierstra,
  D., Riedmiller, M.A.: Playing atari with deep reinforcement learning.
\newblock CoRR \textbf{abs/1312.5602} (2013).
\newblock \urlprefix\url{http://arxiv.org/abs/1312.5602}

\bibitem{mnih2015human}
Mnih, V., Kavukcuoglu, K., Silver, D., Rusu, A.A., Veness, J., Bellemare, M.G.,
  Graves, A., Riedmiller, M., Fidjeland, A.K., Ostrovski, G., et~al.:
  Human-level control through deep reinforcement learning.
\newblock Nature \textbf{518}(7540), 529--533 (2015)

\bibitem{osband2016deep}
Osband, I., Blundell, C., Pritzel, A., Van~Roy, B.: Deep exploration via
  bootstrapped dqn.
\newblock In: Advances in Neural Information Processing Systems, pp. 4026--4034
  (2016)

\bibitem{pavlov}
{Pavlov}, M., {Kolesnikov}, S., {Plis}, S.M.: {Run, skeleton, run: skeletal
  model in a physics-based simulation}.
\newblock ArXiv e-prints  (2017)

\bibitem{plappert}
Plappert, M.: keras-rl.
\newblock \url{https://github.com/matthiasplappert/keras-rl} (2016)

\bibitem{plappert2017parameter}
Plappert, M., Houthooft, R., Dhariwal, P., Sidor, S., Chen, R.Y., Chen, X.,
  Asfour, T., Abbeel, P., Andrychowicz, M.: Parameter space noise for
  exploration.
\newblock arXiv preprint arXiv:1706.01905 (2) (2017)

\bibitem{Rumelhart:1986:LIR:104279.104293}
Rumelhart, D.E., Hinton, G.E., Williams, R.J.: Parallel distributed processing:
  Explorations in the microstructure of cognition, vol. 1.
\newblock chap. Learning Internal Representations by Error Propagation, pp.
  318--362. MIT Press, Cambridge, MA, USA (1986).
\newblock \urlprefix\url{http://dl.acm.org/citation.cfm?id=104279.104293}

\bibitem{progressive}
Rusu, A.A., Rabinowitz, N.C., Desjardins, G., Soyer, H., Kirkpatrick, J.,
  Kavukcuoglu, K., Pascanu, R., Hadsell, R.: Progressive neural networks.
\newblock CoRR \textbf{abs/1606.04671} (2016).
\newblock \urlprefix\url{http://arxiv.org/abs/1606.04671}

\bibitem{salimans}
{Salimans}, T., {Ho}, J., {Chen}, X., {Sidor}, S., {Sutskever}, I.: {Evolution
  Strategies as a Scalable Alternative to Reinforcement Learning}.
\newblock ArXiv e-prints  (2017)

\bibitem{starter}
{Salimans}, T., {Ho}, J., {Chen}, X., {Sidor}, S., {Sutskever}, I.: Starter
  code for evolution strategies.
\newblock \url{https://github.com/openai/evolution-strategies-starter} (2017)

\bibitem{Schaal2006}
Schaal, S.: Dynamic movement primitives -a framework for motor control in
  humans and humanoid robotics.
\newblock In: H.~Kimura, K.~Tsuchiya, A.~Ishiguro, H.~Witte (eds.) Adaptive
  Motion of Animals and Machines, pp. 261--280. Springer Tokyo, Tokyo (2006).
\newblock \doi{10.1007/4-431-31381-8_23}.
\newblock \urlprefix\url{https://doi.org/10.1007/4-431-31381-8_23}

\bibitem{schaul2015prioritized}
Schaul, T., Quan, J., Antonoglou, I., Silver, D.: Prioritized experience
  replay.
\newblock arXiv preprint arXiv:1511.05952  (2015)

\bibitem{ppo}
Schulman, J., Wolski, F., Dhariwal, P., Radford, A., Klimov, O.: Proximal
  policy optimization algorithms.
\newblock CoRR \textbf{abs/1707.06347} (2017).
\newblock \urlprefix\url{http://arxiv.org/abs/1707.06347}

\bibitem{schulman}
{Schulman}, J., {Wolski}, F., {Dhariwal}, P., {Radford}, A., {Klimov}, O.:
  {Proximal Policy Optimization Algorithms}.
\newblock ArXiv e-prints  (2017)

\bibitem{silver2014deterministic}
Silver, D., Lever, G., Heess, N., Degris, T., Wierstra, D., Riedmiller, M.:
  Deterministic policy gradient algorithms.
\newblock In: Proceedings of the 31st International Conference on Machine
  Learning (ICML-14), pp. 387--395 (2014)

\bibitem{stelmaszczyk}
Stelmaszczyk, A., Jarosik, P.: {Our NIPS 2017: Learning to Run source code}.
\newblock \url{https://github.com/AdamStelmaszczyk/learning2run} (2017)

\bibitem{Sutton1999}
Sutton, R.S., Precup, D., Singh, S.: {Between {MDP}s and semi-{MDP}s: A
  framework for temporal abstraction in reinforcement learning}.
\newblock Artificial Intelligence \textbf{112} (1999)

\bibitem{uhlenbeck1930theory}
Uhlenbeck, G.E., Ornstein, L.S.: On the theory of the brownian motion.
\newblock Physical review \textbf{36}(5), 823 (1930)

\bibitem{hq}
Wiering, M., Schmidhuber, J.: {HQ}-learning.
\newblock Adaptive Behaviour \textbf{6} (1997)

\end{thebibliography}
\end{document}